\documentclass[journal]{IEEEtran}
\usepackage{graphicx}
\usepackage{subfigure}

\usepackage{amsmath}   %引入公式包
\usepackage{lineno,hyperref} % 引入宏包
\usepackage{algorithmic}
\usepackage{algorithm}
\usepackage{bbding}
\usepackage{pifont}
\usepackage{wasysym}
\usepackage{amssymb}
\usepackage{color}
\usepackage{amssymb}   
\usepackage{booktabs}   %三线表的宏包
\usepackage{multicol}  
\usepackage{multirow}  
\usepackage{caption}
\usepackage{graphicx}
\usepackage{diagbox}    %插表格用的宏包
\usepackage{multirow}   %插多行表格用的宏包
\usepackage{bm}  %用于公式加粗  %%\textbf{}用于加粗文本

\renewcommand\thetable{\Roman{table}}
\renewcommand{\tablename}{Table}
\modulolinenumbers[5]

\begin{document}
	
\title{Graph-Collaborated Auto-Encoder Hashing for Multi-view Binary Clustering}
\author{Huibing Wang, Mingze Yao, Guangqi Jiang, Zetian Mi, Xianping Fu
	\thanks{H. Wang, M. Yao, M. Ze and X. Fu are with College of Information Science and Technology, Dalian Maritime University, Liaoning, 116026, China, e-mail: (huibing.wang@dlmu.edu.cn; ymz0284@dlmu.edu.cn; mizetian@dlmu.edu.cn; fxp@dlmu.edu.cn). Both Huibing Wang and Mingze Yao are first authors.}
	\thanks{G. Jiang is with School of Computer Science and Artifical Intelligence, Changzhou University, Jiangsu, 213164, China, e-mail: (guangqijiang@cczu.edu.cn). }
\thanks{Mingze Yao, Guangqi Jiang and Xianping Fu are corresponding authors.}}
	
\markboth{}%
{Shell \MakeLowercase{\textit{et al.}}: Bare Demo of IEEEtran.cls for IEEE Journals}

\maketitle

\begin{abstract}
	Unsupervised hashing methods have attracted widespread attention with the explosive growth of large-scale data, which can greatly reduce storage and computation by learning compact binary codes. Existing unsupervised hashing methods attempt to exploit the valuable information from samples, which fails to take the local geometric structure of unlabeled samples into consideration. Moreover, hashing based on auto-encoders aims to minimize the reconstruction loss between the input data and binary codes, which ignores the potential consistency and complementarity of multiple sources data. To address the above issues, we propose a hashing algorithm based on auto-encoders for multi-view binary clustering, which dynamically learns affinity graphs with low-rank constraints and adopts collaboratively learning between auto-encoders and affinity graphs to learn a unified binary code, called Graph-Collaborated Auto-Encoder Hashing for Multi-view Binary Clustering (GCAE). Specifically, we propose a multi-view affinity graphs learning model with low-rank constraint, which can mine the underlying geometric information from multi-view data. Then, we design an encoder-decoder paradigm to collaborate the multiple affinity graphs, which can  learn a unified binary code effectively. Notably, we impose the decorrelation and code balance constraints on binary codes to reduce the quantization errors. Finally, we utilize an alternating iterative optimization scheme to obtain the multi-view clustering results. Extensive experimental results on $5$ public datasets are provided to reveal the effectiveness of the algorithm and its superior performance over other state-of-the-art alternatives.
\end{abstract}

\begin{IEEEkeywords}
 Graph-collaborated, Auto-encoder, Multi-view clustering, Binary code
\end{IEEEkeywords}

\IEEEpeerreviewmaketitle

\section{Introduction}
\IEEEPARstart{W}{ith} the development of information digitization \cite{zhang2022autoencoder, zhang2021dual,wang2018multiview} and computer technology, researchers have proposed a large number of feature extraction methods to extract features from multiple views of the same sample \cite{qian2022switchable}, \cite{wang2021survey}, \cite{peng2020deep}, \cite{wang2015robust}. For example, an image can be extracted as different feature representations by multiple descriptors, i.e., LBP \cite{guo2010completed}, Gabor \cite{movellan2002tutorial}, HOG \cite{dalal2005histograms} and SIFT \cite{lowe2004distinctive}. However, these multi-view data extracted from different feature descriptors have properties that are large-scale and heterogeneous, which cry out for reliable mining methods to explore the discriminative information from multiple views. In order to effectively process the large-scale data, most existing researches introduce hash methods due to its fast running speed and economical storage cost. Specifically, hash methods encode the large-scale data by a set of compact binary codes in a low-dimensional Hamming space. Therefore, existing hash algorithms have been widely applied to various large-scale visual application tasks, such as cross-modal retrieval \cite{gu2019clustering}, object re-identification \cite{wang2022progressive}, image detection \cite{hu2018deep} and multi-view learning \cite{wang2022towards,jiang2022tensorial,wang2020kernelized,wang2018beyond} etc.

Considering the effectiveness of binary codes for various vision tasks with large-scale data, several methods have been proposed to explore the more discriminative binary code representation. Over the past few decades, several supervised hashing methods have been proposed, such as Supervised Discrete Hashing (SDH) \cite{shen2015supervised}, Strongly Constrained Discrete Hashing (SCDH) \cite{chen2020strongly} and Fast Discriminative Discrete Hashing (FDDH) \cite{liu2021fddh}. Note that while these aforementioned approaches have achieved great performance with hashing, most of them deeply depend on the manual labels, which is time-consuming and less effective process the large-scale unlabeled data. Therefore, some unsupervised hashing methods have been proposed to deal with the unlabeled problem. The typical unsupervised hashing is Locality-Sensitive Hashing (LSH) \cite{indyk1998approximate} which adopts random projections to generate discrete binary codes. Based on LSH, Spectral Hashing (SH) \cite{weiss2008spectral}, Discrete Graph Hashing (DGH) \cite{liu2014discrete} and Scalable Graph Hashing (SGH) \cite{jiang2015scalable} have been proposed to explore similar information from the large-scale data. Even though the above methods have effectively learned compact binary codes in an unsupervised manner, most existing hashing methods usually utilize the data from single source. For multi-view data, these hashing methods are difficult to uncover the multi-view information holistically and ignore the consistent and complementary information from different views.

Compared with the data from a single source, multi-view data usually contain more compatible and complementary information hidden in different views, which are extracted from same samples. Therefore, multi-view clustering methods have been proposed to explore the latent structure of different views and integrate complementary information from multi-view data. Kumar et al. \cite{kumar2011co} introduced a co-regularized model to complete spectral clustering with a centroid-based algorithm and pairwise algorithm which can mine the underlying structure from original data. Zhan et al. \cite{8052206} proposed a graph-learning method with the rank constraint to integrate different graphs into a global graph for multi-view clustering tasks. Wang et al. \cite{wang2016iterative} proposed a multi-graph laplacian regularized LRR model, which can separately impose a low-rank constraint on each graph to achieve agreeable results. Besides, Wang et al. \cite{8662703} performed reinforcement learning on the graph of each view and the unified graph of all views by considering the weights of different views. Xiao et al. \cite{8778709} proposed a graph-based multi-view clustering framework with knowledge elements, which can combine knowledge and language for clustering. Moreover, Shi et al. \cite{9492299} proposed a common joint graph learning strategy, which utilizes nonnegative constraint to fully explore the structure information from multi-view data. This strategy aims to directly obtain cluster results and avoid post-processing. The above methods mostly measure the distance between features in Euclidean space, while they still need a high computational cost and low efficiency for processing large-scale data.
\begin{figure*}[tbp!]
	\centering
	\includegraphics[width=\textwidth]{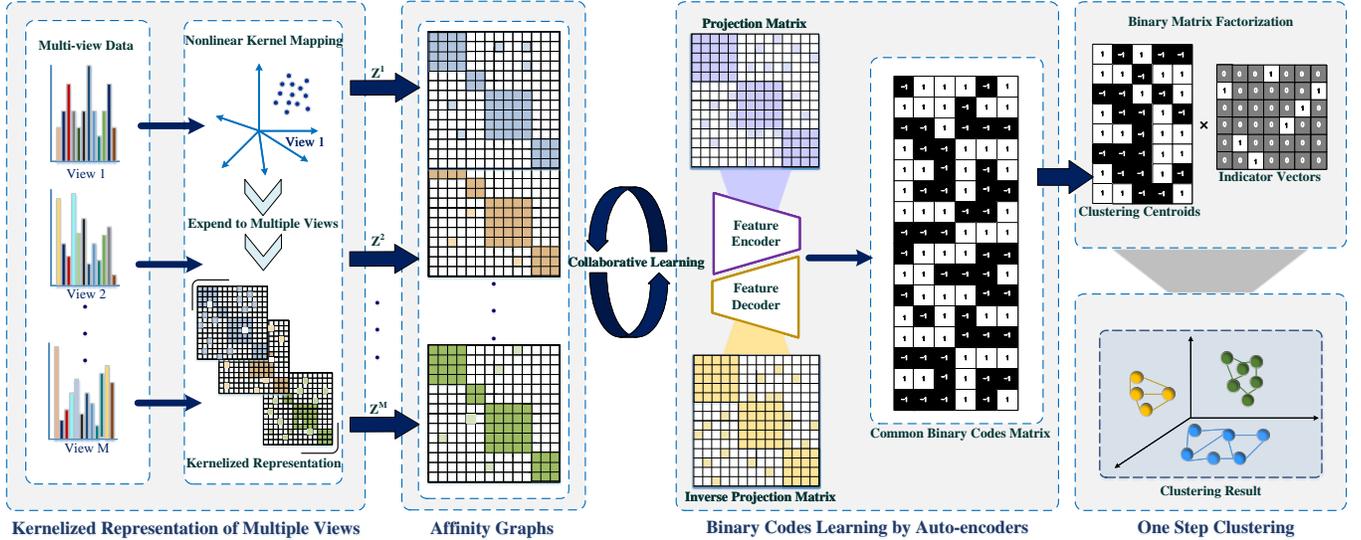} 
	\caption{The whole model of Graph-Collaborated Auto-Encoder Hashing for Multi-view Binary Clustering (GCAE), which learns affinity graphs by low-rank constraint to integrate specific information from multiple views and adopts auto-encoder to generate unified binary code for clustering. }
	\label{whole}
\end{figure*}

Some researchers proposed multi-view hash methods to learn compact binary codes and utilize efficient XOR operation \cite{zhang2018binary}, which can improve the speed and accuracy of the algorithm. Jin et al. \cite{6406693} proposed a binary function clustering scheme that captures the function semantics as semantic hashing to quickly cluster the high degree of similarity samples. Tian et al. \cite{tian2020unsupervised} provided a variant of the LRR \cite{liu2010robust} model to recover the latent structure of original data, which can effectively learn similarity graphs for binary code learning. Wang et al. \cite{wang2017robust} utilized $l_{2,1}$-norm to learn compact binary codes, which can improve the robustness of the model. Shen et al. \cite{wang2021set} constructed a novel semantic-rebased model, which adopted a sparse graph setting and rebased the similarity graph. Notably, most related hashing works focus on retrieval tasks, which ignore the complementary information and underlying cluster structure from multi-view data. Recently, several hashing algorithms have been proposed to solve large-scale image clustering problems. Wang et al. \cite{wang2021cluster} provided a cluster-wise unsupervised hashing framework, which projects the multi-view original data into latent low-dimensional space to learn cluster centroid for searching. Zhang et al. \cite{zhang2018highly} explored a highly-economized algorithm for image clustering, which jointly learned binary representation and binary cluster results. Even though the above methods can process large-scale data effectively and achieve great performance, most of them heavily rely on affinity graphs from original data directly and fail to mine the local structures. Meanwhile, some studies simplified the optimization problem by relaxing binary constraints, which may cause quantization errors. Therefore, it is essential to compose an effective graph collaboration framework to explore the local geometric information from multiple views and utilize suitable binary constraints.

To address the above limitations, this paper proposes a novel method, termed as Graph-Collaborated Auto-Encoder Hashing for Multi-view Binary Clustering (GCAE). GCAE constructs auto-encoders to learn binary codes for processing multi-view data, which emphasizes collaboratively learning between affinity graphs and auto-encoders to learn a unified binary codes for multi-view clustering. Firstly, GCAE constructs affinity graphs from each view by imposing a low-rank constraint on the original data, which can preserve essential information and the latent structure from the multi-view data. Secondly, to effectively explore the compatible and complementary information from multi-view data, GCAE adopts auto-encoders to collaborate multiple affinity graphs, which aim to learn unified binary codes for clustering and preserve the discrete binary constraint. Subsequently, GCAE utilizes the matrix factorization strategy to directly obtain cluster results without post-processing, which can avoid error accumulation. Finally, an alternating iterative optimization strategy is adopted to update each variable of the objective function. The whole model of GCAE has been shown in Fig. \ref{whole}. The major contributions of the proposed method are summarized as follows:

\begin{itemize}
	\item We propose Graph-Collaborated Auto-Encoder Hashing for Multi-view Binary Clustering (GCAE), which utilizes affinity graphs and auto-encoders collaboratively to learn compact binary codes for multi-view clustering.
	\item In particular, GCAE imposes the low-rank constraint on graphs to mine essential information effectively and utilizes auto-encoders to collaborate multiple graphs for learning unified binary codes, which can explore complementary information from multi-view data and guide the learning of binary codes. Besides, our proposed  GCAE directly obtain cluster results to avoid the accumulation of errors caused by post-processing.
\end{itemize}

The remainder of the paper is outlined as follows. Section 2 introduces the related work. Section 3 presents the proposed GCAE model and the optimization process. Extensive experiments including complexity analysis and convergence analysis are conducted to verify our proposed model in Section 4. Finally, Section 5 concludes this paper.

\section{Related Work}
In this section, we briefly review the related studies about graph-based multi-view clustering and multi-view hashing methods with graphs.

Graph-based multi-view clustering methods mostly aim to integrate information from multiple views and calculate similarity graphs in Euclidean distance for clustering. For example, Nie et al. \cite{nie2016parameter} proposed a framework based on standard spectral learning which learns weights for multiple graphs automatically without introducing additive parameters. Hou et al. \cite{7876761} presented an automatic method to learn a common similarity graph to characterize the structures across different views and tune balance weights.  However, the above methods require an additional clustering step to obtains the final clusters by utilizing K-means \cite{hartigan1979algorithm}. In order to avoid the impact of post-processing for obtain the cluster results, Wang et al. \cite{8662703} proposed a model which can produce clusters directly without post-processing for clustering and construct each view graph and fusion graph simultaneously. Besides, Zhang et al. \cite{zhan2018graph} utilize Hadamard product to integrate multiple graphs into a global graph which can recover the graph structure effectively. Shi et al. \cite{shi2020auto} proposed a unified framework for jointly learning multiple similarity graphs and spectral embedding, which can obtain cluster results in a unified framework. Although the above methods achieved great clustering results, they directly build graphs by original data, which may make the learned similarity graph inaccurate and ignore the underlying cluster structure of multi-view data. Meanwhile, the real-value multi-view data requires high computational costs for generating graphs with large-scale multi-view data. Some researchers explore hash methods which utilize binary codes rather than real-value features. Therefore, multi-view hashing methods are widely used to integrate large-scale data with the recent advances.

Existing multi-view hashing methods with graphs can be roughly divided into supervised and unsupervised methods based on the usage of semantic presentation (e.g., class labels). For supervised methods, Jin et al. \cite{8118130} construct a semantic graph by jointly taking the semantic presentation and the local similarity structure into consideration. Guan et al. \cite{guan2019graph} proposed a supervised learning model to construct similarity graphs that capture the intrinsic manifold structure from semantic supervision. Despite these supervised methods can achieve great performance, it is not practical to label information manually for large-scale data. Therefore, Liu et al. \cite{liu2014discrete} utilized anchor graphs to capture the latent structure inherent in a given massive dataset and calculated hash codes in Hamming space. Xiang et al. \cite{xiang2019discrete} proposed a novel hashing method that adopted a quantization regularization term to reduce the distortion error during constructing similarity graphs. Besides, Fang et al. \cite{fang2019unsupervised} jointly learned the intra-modal similarity graph and reconstructed the similarity graph by symmetric nonnegative matrix factorization and then utilized binary code inner product to learn binary codes. The above approaches have obtained good results, but relaxing the discrete constraint on binary codes will cause quantization errors. Meanwhile, for the multi-view clustering tasks, existing multi-view hashing methods have not effectively mined the underlying cluster structure from different views. 

\section{Graph-Collaborated Auto-Encoder Hashing}
In this paper, boldface lowercase letters and boldface uppercase letters are used to denote the vectors and matrices, receptively, e.g., $\bm{{\rm{x}}}$ and $\bm{{\rm{X}}}$. For any matrix $\bm{{\rm{X}}}$, $\bm{{\rm{x}}}_{ij}$ means the $(i,j)$-element of $\bm{{\rm{X}}}$. Besides some important mathematic notations are listed in Table \ref{Table1}. For given multi-data sets, $\bm{{\rm{X}}}=\{\bm{{\rm{X}}}^1,\bm{{\rm{X}}}^2,...,\bm{{\rm{X}}}^v\}$ where ${\bm{{\rm{X}}}^v} \in {\mathbb{R}} ^{N \times {d^v}}$ is the $v$-th view data. ${d^v}$ denotes the feature dimension and $N$ is the number of sample.
\begin{table}[tbp!]
	\centering
	\Large
	\caption{The Descriptions of Some Important Formula Symbols.} 
	\label{Table1}
	\scalebox{0.7}{\begin{tabular}{c|c|c}
			\hline
			Scalar & $N$                   & $N \in {\mathbb{R}} $              \\ \hline
			Vector &  $\bm{{\rm{a}}}$             &       $\bm{{\rm{a}}} \in {{\mathbb{R}}^N}$                           \\
			& $\bm{{\rm{a}}}_i$                      &     $i$-th row                             \\
			&  $\bm{{\rm{a}}}_i^{v}$                  &    $i$-th  row with $v$-th view                         \\
			&         $\bm{1}$              &             $all$-1 vector                     \\
			&         $\bm{0}$           &      $all$-0 vector \\
			&       $||\bm{{\rm{a}}}||_2$   &  $l_2$ norm as $\sqrt {\sum {_{i = 1}^N\bm{{\rm{a}}}_i^2} } $  \\ \hline
			Martix & $\bm{{\rm{A}}}^v$            & $\bm{{\rm{A}}}^v \in {\mathbb{R}} ^{N \times {d^v}}$   \\
			&     $\bm{{\rm{A}}}_{i,j}$       &      $(i,j)$-th element      \\       
			& $\bm{{\rm{I}}}$              & the identity matrix                  \\
			& $tr(\bm{{\rm{A}}})$          &  the trace of matrix $\bm{{\rm{A}}}$     \\
		    & $\bm{{\rm{A}}}^T$            &  the transpose of matrix $\bm{{\rm{A}}}$ \\ 
		    & $\sigma$     &$\sigma  = diag(\bm{{\rm{S}}})$ \\
		    	& $||\bm{{\rm{A}}}||_F $       &  Frobenius norm as$ \sqrt {\sum {_{i,j}\bm{{\rm{A}}}_{i,j}^2,{{\left\| \sigma  \right\|}_2}} }  $ \\
		    & $||\bm{{\rm{A}}}||_* $       &  nuclear norm as $tr(\sqrt {{\bm{{\rm{A}}}^T}\bm{{\rm{A}}}} )$    \\
		    & ${\phi (\bm{{\rm{A}}}^v)}$ &  kernelized representation of $(\bm{{\rm{A}}}^v)$          \\\hline
		\end{tabular}
		}
\end{table}

We propose a novel model of graph-collaborated auto-encoder hashing for clustering. Specifically, we jointly learned the affinity graphs from each view and constructed auto-encoders for the unified hash code learning which can minimize reconstruction loss between affinity graphs and hash code. Besides, our method utilized matrix factorization strategy to obtain cluster results.
\subsection{Multi-view Affinity Graph Learning}
For given data matrix $\bm{{\rm{X}}}=\{\bm{{\rm{X}}}^1,\bm{{\rm{X}}}^2,...,\bm{{\rm{X}}}^v\}$, where $\bm{{\rm{X}}}^v$ presents the $v$-th view data matrix. Different from existing methods that directly utilize original multi-view data for generating multiple graphs. GCAE firstly adopts nonlinear kernel mapping for each view, which aims to unify the dimension of multi-view data so that we can learn the underlying geometric structure from data. Inspired by \cite{zhang2018binary}, we employ the nonlinear RBF kernel mapping method for each view as follows:
\begin{gather}
\label{1}
\phi (\bm{{{\rm{X}}}}^v) = {[\exp ( - {\left\| {\bm{{\rm{x}}}_1^v - \bm{{\rm{a}}}_1^v} \right\|^2}/\eta ),...,\exp ( - {\left\| {\bm{{\rm{x}}}_N^v - \bm{{\rm{a}}}_t^v} \right\|^2}/\eta )]^T}
\end{gather}
where $\eta$ is the kernel width, $\phi ({{\bm{{{\rm{X}}}}}^v}) \in {\mathbb{R}} ^{N\times t}$ indicates the $t$-dimensional nonlinear mapping from the $v$-th view. Besides, to ensure that the original data structure is not destroyed after nonlinear projection, we randomly select $t$ anchor samples $\bm{a}_t^v$ from the $v$-th view.

After $\phi ({{\bm{{{\rm{X}}}}}^v})$ yielded by Eq. \ref{1}, the affinity graph $\bm{{\rm{Z}}}^v$ is constructed by processing each view of $\phi ({{\bm{{{\rm{X}}}}}^v})$. Above this, we propose a new variant of the LRR \cite{li2015learning} model that can make the learned affinity graph retain the important information of the original data instead of noise and redundant information. Then, we give the following model:
\begin{gather}
\label{2}
\mathop {\min }\limits_{\bm{{\rm{Z}}}^v} \sum\limits_{v = 1}^M {\left\| {\phi ({\bm{{\rm{X}}}^v}) - {\bm{{\rm{Z}}}^v}\phi ({\bm{{\rm{X}}}^v})} \right\|_F^2 + {{\left\| {{\bm{{\rm{Z}}}^v}} \right\|}_*}} 
\end{gather}
where ${\bm{{\rm{Z}}}^v} \in {{\mathbb{R}} ^{N \times N}}$ represents the learned affinity graph for $v$-th view data. $||\bm{{\rm{Z}}}^v||_* $ indicates the nuclear norm which calculates the sum of singular values for low-rank constraint. Minimizing the objective function in Eq. \ref{2} aims to preserve important information while exploring the low-rank structure from the original data. 
\subsection{Graph-collaborated Auto-Encoders Hashing for Clustering }
In order to encourage collaborative learning of affinity graphs from different views, we proposed an auto-encoders hashing by the multi-graphs model which can effectively mine the consistency and complementarity information and generate binary code for clustering. Different from the existing hash methods, the auto-encoders hashing model preserves the discrete constraint and the uncorrelated constraint on binary bits.
\begin{gather}
\begin{aligned}	
\label{3}
&\mathop {\min }\limits_{\bm{{\rm{W}}}^v,\bm{{\rm{B}}},\bm{{\rm{Z}}}^v} \sum\limits_{v = 1}^M ( \left\| {{\bm{{\rm{W}}}^v}{\bm{{\rm{Z}}}^v} - \bm{{\rm{B}}}} \right\|_F^2 + \| {{\bm{{\rm{Z}}}^v} - {\bm{{\rm{W}}}^{vT}}\bm{{\rm{B}}}} \|_F^2)\\
&s.t.\quad {\bm{{\rm{W}}}^v}{\bm{{\rm{W}}}^v}^T = \bm{{\rm{I}}},\bm{{\rm{B}}} \in {\{  - 1, + 1\} ^{b \times N}},\bm{{\rm{B}}}{\bm{{\rm{B}}}^T} = N\bm{{\rm{I}}}
\end{aligned}
\end{gather}

Eq. \ref{3} aims to minimize the difference between low-rank affinity graphs mapping and discrete hash code. $b$ represents binary bits. The decorrelation and code balance constraints $\bm{{\rm{B}}} \in {\{  - 1, + 1\} ^{b \times N}},\bm{{\rm{B}}}{\bm{{\rm{B}}}^T} = N\bm{{\rm{I}}}$ are imposed to generate mutually independent binary codes and reduce quantization errors. $\bm{{\rm{W}}}^v$ presents the projection matrix. For simplicity, we utilize the transpose matrix of $\bm{{\rm{W}}}^v$ to replace the inverse mapping matrix in auto-encoder, and add a constraint $\bm{{\rm{W}}}^v\bm{{\rm{W}}}^{vT}=\bm{{\rm{I}}}$ on projection matrix. Meanwhile, the regularization term $||\bm{{\rm{W}}}^v||_F^2$ is the constant due to $||\bm{{\rm{W}}}^v||_F^2=tr(\bm{{\rm{W}}}^v\bm{{\rm{W}}}^{vT})=tr(\bm{{\rm{I}}})=const$. Finally, we construct a matrix factorization model with the generated unified binary codes, which can avoid the suboptimal results caused by the two-step clustering methods. Above all, Eq. \ref{3} can be reformulated as follows:
\begin{gather}
\label{4}
\begin{aligned}
&\mathop {\min }\limits_{{\bm{{\rm{Z}}}^v},{\bm{{\rm{W}}}^v},\bm{{\rm{B}}},\bm{{\rm{Q}}},\bm{{\rm{H}}}} \sum\limits_{v = 1}^M  (\left\| {{\bm{{\rm{W}}}^v}{\bm{{\rm{Z}}}^v} - \bm{{\rm{B}}}} \right\|_F^2+ \| {{\bm{{\rm{Z}}}^v} - {\bm{{\rm{W}}}^{vT}}\bm{{\rm{B}}}} \|_F^2)) \\&\qquad\qquad\qquad+  \lambda \left\| {\bm{{\rm{B}}} - \bm{{\rm{QH}}}} \right\|_F^2\\
&s.t.\quad{\bm{{\rm{W}}}^v}{\bm{{\rm{W}}}^{vT}} = \bm{{\rm{I}}},\bm{{\rm{B}}} \in \left\{ { - 1,} \right.{\left. { + 1} \right\}^{b \times N}},\bm{{\rm{B}}}{\bm{{\rm{B}}}^T} = N\bm{{\rm{I}}},\\& {\bm{{\rm{Q}}}^T}\bm{{\rm{1}}} = \bm{{\rm{0}}}, \bm{{\rm{Q}}} \in \left\{ { - 1,} \right.{\left. { + 1} \right\}^{b \times c}},\bm{{\rm{H}}}\in \left\{ {0,} \right.{\left. 1 \right\}^{c \times N}},\sum\limits_{i=1}^c {{\bm{{\rm{h}}}_{is}} = 1} 
\end{aligned}
\end{gather}
where $\lambda$ is the regularization parameter and $c$ is the number of clusters. $\bm{{\rm{Q}}}$ and $\bm{{\rm{H}}}$ represent the clustering centroids and cluster indicator matrix, respectively. Meanwhile, in order to generate the efficient binary code and maximize the information of each bits for clustering, we add the balance constraint on clustering centroids $\bm{{\rm{Q}}}$, which can produce efficient code and make our model to adapt binary clustering task. 
\subsection{Overall Objective Function of GCAE}
In order to collaborating affinity graphs and auto-encoders to learn a unified binary code  for clustering, we summarize the above graph-collaborated auto-encoders hashing and the low-rank affinity graphs learning, which are both crucial to multi-view clustering. The above consideration can be fulfilled as follows:
\begin{gather}
\begin{aligned}
\label{5}
&\min {\cal L}({\bm{{\rm{Z}}}^v},{\bm{{\rm{W}}}^v},\bm{{\rm{B}}},\bm{{\rm{Q}}},\bm{{\rm{H}}},{\bm{{\rm{p}}}^v})\\ &= \underbrace {\sum\limits_{v = 1}^M {\left\| {\phi ({\bm{{\rm{X}}}^v}) - {\bm{{\rm{Z}}}^v}\phi ({\bm{{\rm{X}}}^v})} \right\|_F^2 + {{\left\| {{\bm{{\rm{Z}}}^v}} \right\|}_*}} }_{{\rm{Multi - view \hspace{1mm}  Affinity\hspace{1mm}  Graph\hspace{1mm} Learning}}} +\\&\!\underbrace {\sum\limits_{v = 1}^M\! {{{({\bm{{\rm{p}}}^v})}^k}\!(\left\| {{\bm{{\rm{W}}}^v}\!{\bm{{\rm{Z}}}^v} \!-\! \bm{{\rm{B}}}} \right\|_F^2\! + \!\| {{\bm{{\rm{Z}}}^v}\! -\! {\bm{{\rm{W}}}^v}^T\!\bm{{\rm{B}}}} \|_F^2)}  \!+\! \lambda\! \left\| {\bm{{\rm{B}}}\! - \!\bm{{\rm{Q\!H}}}} \right\|_F^2}_{{\rm{Auto - Encoders \hspace{1mm} Hashing\hspace{1mm}  by\hspace{1mm} Multi - Graphs\hspace{1mm} for\hspace{1mm} Clustering }}}\\
&s.t.\quad {\bm{{\rm{W}}}^v}{\bm{{\rm{W}}}^{vT}} = \bm{{\rm{I}}},\bm{{\rm{B}}} \in \left\{ { - 1,} \right.{\left. { + 1} \right\}^{b \times N}},\bm{{\rm{B}}}{\bm{{\rm{B}}}^T} = N\bm{{\rm{I}}},\\&\qquad{\bm{{\rm{Q}}}^T}{\bm{{\rm{1}}}} = \bm{{\rm{0}}},\bm{{\rm{Q}}} \in \left\{ { - 1,} \right.{\left. { + 1} \right\}^{b \times c}},\bm{{\rm{H}}} \in \left\{ {0,} \right.{\left. 1 \right\}^{c \times N}},\\&\qquad\sum\limits_{v=1}^M {{\bm{{\rm{p}}}^v} = 1} ,{\bm{{\rm{p}}}^v} > 0,\sum\limits_{i=1}^c {{\bm{{\rm{h}}}_{is}} = 1} 
\end{aligned}
\end{gather}
where ${\bm{{\rm{p}}}^v}$ indicates the normalized weighting coefficient, which aims to balance the affinity graphs from each view according to the contribution. Besides. we also add the constraint ${\bm{{\rm{p}}}^v>0}$ to ensure nonnegative vector. Above all, we summarize multi-view affinity graph learning and auto-encoders hashing by multi-graphs in a unified framework. The proposed GCAE model learns the low-rank affinity graph from each view, which can effectively preserve important information from original data to improve the quality of affinity graphs. And then, the proposed model utilizes binary matrix factorization model for the unified binary codes, which can effectively generate cluster results in a one-step model. By jointly optimizing the above equation, we obtain the graph-collaborated binary code representation to obtain the cluster results. We propose a novel optimization algorithm for the objective function Eq. \ref{5} in the following section.
\subsection{Optimization Process for GCAE}
In this section, we describe the optimization process for GCAE in detail. It is obvious that Eq. \ref{5} is a nonconvex optimization problem that can not directly obtain the holistic optimal solution in Eq. \ref{5}. We develop an auxiliary matrices strategy, which separates the problem into two sub-problems include low-rank affinity graphs learning and auto-encoders hash for obtaining optimized cluster results. Based on the auxiliary matrices strategy, we introduce two auxiliary matrices to relax the nuclear norm and control the rank of affinity graphs. And then the proposed auto-encoders utilize the affinity graphs from each view to generate binary code for clustering. Finally, we develop an iterative alternative strategy, which alternatively updates each variable when fixing others. 

The proposed auxiliary matrices strategy introduces the multiplication of two auxiliary matrices $\bm{{\rm{F}}}^v$ and $\bm{{\rm{G}}}^v$ (i.e., $\bm{{\rm{Z}}}^v=\bm{{\rm{F}}}^v\bm{{\rm{G}}}^{vT}$) to replace the low-rank affinity graph $\bm{{\rm{Z}}}^v$, and then problem Eq. \ref{2} can be converted as:
\begin{equation}
\label{6}
\mathop {\min }\limits_{\bm{{\rm{F}}}^v,\bm{{\rm{G}}}^v} \left\| {\phi ({\bm{{\rm{X}}}^v}) - {\bm{{\rm{F}}}^v}\bm{{\rm{G}}}^{vT}}\phi ({\bm{{\rm{X}}}}^v) \right\|_F^2
\end{equation}
where $\bm{{\rm{F}}}^v \in { {\mathbb{R}} ^{N \times r}}$ and $\bm{{\rm{G}}}^v \in { {\mathbb{R}} ^{N \times r}}$. $r$ is a parameter that we use to relax nuclear norm and control the rank of $\bm{{\rm{Z}}}^v$. And this strategy is based on the fact that $rank({\bm{{\rm{Z}}}^v}) = rank({\bm{{\rm{F}}}^v}{\bm{{\rm{G}}}^{vT}}) \le \min (rank({\bm{{\rm{F}}}^v}),rank({\bm{{\rm{G}}}^v})) \le r$ (generally $r\ll N$). The updating process of $\bm{{\rm{F}}}^v$ and $\bm{{\rm{G}}}^v$ are shown as follows:

$\bm{{\rm{Updating}} }\quad \bm{{\rm{F}}}^v$: Based on the operational rules of matrix trace, Eq. \ref{6} can be unfolded as:
\begin{gather}
\begin{aligned}
&O(\bm{{\rm{F}}}^v) =\\ &tr((\phi ({\bm{{\rm{X}}}^v}) - {\bm{{\rm{F}}}^v}{\bm{{\rm{G}}}^{vT}}\phi ({\bm{{\rm{X}}}^v})){(\phi ({\bm{{\rm{X}}}^v}) - {\bm{{\rm{F}}}^v}{\bm{{\rm{G}}}^{vT}}\phi ({\bm{{\rm{X}}}^v}))^T}) 
\end{aligned}
\end{gather}
then we calculate $\bm{{\rm{F}}}^v$ iteratively by setting the derivation $\frac{{\partial O({\bm{{\rm{F}}}^v})}}{{\partial {\bm{{\rm{F}}}^v}}} = \bm{{\rm{0}}}$:
\begin{gather}
\begin{aligned}
\frac{{\partial O({\bm{{\rm{F}}}^v})}}{{\partial {\bm{{\rm{F}}}^v}}} =& 2{\bm{{\rm{F}}}^v}{\bm{{\rm{G}}}^{vT}}\phi({\bm{{\rm{X}}}^v})\phi({\bm{{\rm{X}}}^{v})^T}{\bm{{\rm{G}}}^v}\\& - 2\phi({\bm{{\rm{X}}}^v})\phi({\bm{{\rm{X}}}^{v})^T}{\bm{{\rm{G}}}^v} = \bm{{\rm{0}}}
\end{aligned}
\end{gather}

Obviously, the close solution of $\bm{{\rm{F}}}^v$ is written as:
\begin{gather}
\label{9}
{\bm{{\rm{F}}}^v} = \phi ({\bm{{\rm{X}}}^v})\phi {({\bm{{\rm{X}}}^v)}^T}{\bm{{\rm{G}}}^v}(\bm{{\rm{G}}}^{vT}\phi ({\bm{{\rm{X}}}^v)}\phi {({\bm{{\rm{X}}}^v)}^T}{\bm{{\rm{G}}}^v})^{ - 1}
\end{gather}

$\bm{{\rm{Updating}} }\quad \bm{{\rm{G}}}^v$: With $\bm{{\rm{F}}}^v$ fixed, we can solve the optimization problem of $\bm{{\rm{G}}}^v$ which is similar with above method. The solution of $\bm{{\rm{G}}}^v$ can be easily obtained:
\begin{gather}
\label{10}
\bm{{\rm{G}}}^v = \bm{{\rm{F}}}^v{({\bm{{\rm{F}}}^{vT}}\bm{{\rm{F}}}^v)^{ - 1}}
\end{gather}

Based on the above solution of $\bm{{\rm{F}}}^v$ and $\bm{{\rm{G}}}^v$, we summarize the above iteratively method in Algorithm 1. Notably, in order to avoid obtaining the singular value, we add a small smooth item in optimization. That is, ${\bm{{\rm{F}}}^v} = \phi ({\bm{{\rm{X}}}^v})\phi {({\bm{{\rm{X}}}^v)}^T}{\bm{{\rm{G}}}^v}(\bm{{\rm{G}}}^{vT}\phi ({\bm{{\rm{X}}}^v)}\phi {({\bm{{\rm{X}}}^v)}^T}{\bm{{\rm{G}}}^v}+\theta\bm{{\rm{I}}})^{ - 1}$ and $\bm{{\rm{G}}}^v = \bm{{\rm{F}}}^v{({\bm{{\rm{F}}}^{vT}}\bm{{\rm{F}}}^v+\theta\bm{{\rm{I}}})^{ - 1}}$ where $\theta$ is set in range of $[1e^{-4},1e^{-6}]$ in our paper.	
\begin{algorithm}[htbp!] 
	\caption{Low-Rank Affinity Graphs Learning Algorithm}  
	\label{alg:A}  
	\begin{algorithmic}[1] 
		\REQUIRE {Kernelized dataset $\phi({\bm{{\rm{X}}}^v});$ the parameter $r$, and the maximum number of iterations $Iter$} 
		\ENSURE  $\bm{{\rm{F}}}^v$, $\bm{{\rm{G}}}^v$  
		\STATE set $t=1$ and randomly initialize $\bm{{\rm{F}}}^{(0)v}$ and $\bm{{\rm{G}}}^{(0)v}$ 
		\WHILE {$t<Iter$}
		\STATE Compute $\bm{{\rm{F}}}^{(t)v}$ according to Eq. \ref{9}
		\STATE Compute $\bm{{\rm{G}}}^{(t)v}$ according to Eq. \ref{10}
		\STATE $t=t+1$
		\ENDWHILE   
		\RETURN  $\bm{{\rm{F}}}^{(t)v}$ and $\bm{{\rm{G}}}^{(t)v}$ 
	\end{algorithmic}  
\end{algorithm}  

After the above low-rank affinity graphs learning with auxiliary matrices strategy, we have rewritten the formula to auto-encoders hash for clustering as Eq. \ref{11}. In order to obtain the optimal solution of Eq. \ref{11}, we propose an iterative optimization method. The proposed method can effectively maintain discrete constraints for binary code and get more efficient binary code for clustering.
\begin{gather}
\begin{aligned}
\label{11}
&\mathop {\min}\limits_{\bm{{\rm{Z}}}^v,\bm{{\rm{B,Q,H}}},\bm{{\rm{p}}}^v} \sum\limits_{v = 1}^M {(\| {{\bm{{\rm{F}}}^v}{\bm{{\rm{G}}}^v}^T - {\bm{{\rm{Z}}}^v}} \|_F^2 + {{({\bm{{\rm{p}}} ^v})}^k}} (\left\| {{\bm{{\rm{W}}}^v}{\bm{{\rm{Z}}}^v} - \bm{{\rm{B}}}} \right\|_F^2 \\&\hspace{5em}+ \| {{\bm{{\rm{Z}}}^v} - {\bm{{\rm{W}}}^v}^T\bm{{\rm{B}}}} \|_F^2)) + \lambda \left\| {\bm{{\rm{B}}} - \bm{{\rm{QH}}} } \right\|_F^2\\
&s.t.\quad{\bm{{\rm{W}}}^v}{\bm{{\rm{W}}}^{vT}} = \bm{{\rm{I}}},\bm{{\rm{B}}} \in \left\{ { - 1,} \right.{\left. { + 1} \right\}^{b \times N}},\bm{{\rm{BB}}}^T = N\bm{{\rm{I}}},\\&\qquad\bm{{\rm{Q}}}^T\bm{{\rm{1}}} = \bm{{\rm{0}}},\bm{{\rm{Q}}} \in \left\{ { - 1,} \right.{\left. { + 1} \right\}^{b \times c}}, \bm{{\rm{H}}} \in \left\{ {0,} \right.{\left. 1 \right\}^{c \times N}},\\&\qquad\sum\limits_{v=1}^M {\bm{{\rm{p}}} ^v = 1,} \bm{{\rm{p}}} ^v > 0,\sum\limits_{i=1}^c {{\bm{{\rm{h}}}_{is}} = 1}
\end{aligned}
\end{gather}
$\bm{{\rm{Updating}} }\quad \bm{{\rm{Z}}}^v$: By fixing all variables but $\bm{{\rm{Z}}}^v$, problem(\ref{11}) reduces to:
\begin{gather}
\begin{aligned}
\mathop {\min}\limits_{\bm{{\rm{Z}}}}& \sum\limits_{v = 1}^M (\| {{\bm{{\rm{F}}}^v}{\bm{{\rm{G}}}^{vT}} - {\bm{{\rm{Z}}}^v}} \|_F^2\\& + {{({\bm{{\rm{p}}} ^v})}^k} (\left\| {{\bm{{\rm{W}}}^v}{\bm{{\rm{Z}}}^v} - \bm{{\rm{B}}}} \right\|_F^2 + \| {{\bm{{\rm{Z}}}^v} - {\bm{{\rm{W}}}^{vT}}\bm{{\rm{B}}}} \|_F^2))
\end{aligned}
\label{12}
\end{gather}

In order to minimize the above equation, we convert the Frobenius norm in the above equation to the trace form of the matrix which is convenient for derivation. And then we take the derivative of the trace of the matrix with regard to $\bm{{\rm{Z}}}^v$ as follows:
\begin{gather}
\begin{aligned}
\frac{{\partial {{\cal L}_{{\bm{{\rm{Z}}}^v}}}}}{{\partial {\bm{{\rm{Z}}}^v}}} &= \sum\limits_{v = 1}^M tr({\bm{{\rm{Z}}}^v}{\bm{{\rm{Z}}}^{vT}} - 2{\bm{{\rm{F}}}^v}{\bm{{\rm{G}}}^v}^T{\bm{{\rm{Z}}}^{vT}} \\&+ \!{{({\bm{{\rm{p}}} ^v})}^k}\!({\bm{{\rm{W}}}^v}\!{\bm{{\rm{Z}}}^v}\!{\bm{{\rm{Z}}}^{vT}}\!{\bm{{\rm{W}}}^{vT}} \!+\! {\bm{{\rm{Z}}}^v}\!{\bm{{\rm{Z}}}^{vT}}\! -\! 4{\bm{{\rm{W}}}^{vT}}\!\bm{{\rm{B}}}\!{\bm{{\rm{Z}}}^{vT}}))
\end{aligned}
\end{gather}
whose solution can be easily achieved by setting the derivation to $\bm{{\rm{0}}}$:
\begin{gather}
\begin{aligned}
&{\bm{{\rm{Z}}}^v} = \\&{({({\bm{{\rm{p}}}^v})^k\bm{{\rm{W}}}^{vT}}{\bm{{\rm{W}}}^v} + \bm{{\rm{I}}} + ({\bm{{\rm{l}}}^v})^k\bm{{\rm{I}}})^{ - 1}}({\bm{{\rm{F}}}^v}{\bm{{\rm{G}}}^{vT}} + 2{({\bm{{\rm{l}}} ^v})^k}{\bm{{\rm{W}}}^{vT}}\bm{{\rm{B}}})
\end{aligned}
\end{gather}

$\bm{{\rm{Updating}} }\quad \bm{{\rm{W}}}^v$: In this stage, other variables are fixed. We take the constraint $\bm{{\rm{W}}}^v\bm{{\rm{W}}}^{vT}=\bm{{\rm{I}}}$ in consider, the whole loss function related to $\bm{{\rm{W}}}^v$ in Eq. \ref{11} can be rewritten as:
\begin{gather}
\begin{aligned}
\label{15}
{{\cal L}_{{\bm{{\rm{W}}}^v}}} &= \mathop {\min}\limits_{\bm{{\rm{W}}}^v} \sum\limits_{v = 1}^M ( \left\| {{\bm{{\rm{W}}}^v}{\bm{{\rm{Z}}}^v} - \bm{{\rm{B}}}} \right\|_F^2 + \| {{\bm{{\rm{Z}}}^v} - {\bm{{\rm{W}}}^{vT}}\bm{{\rm{B}}}} \|_F^2) \\
&=\max tr({\bm{{\rm{W}}}^v}{\bm{{\rm{Z}}}^v}{\bm{{\rm{B}}}^T}) \\& s.t.\quad{\bm{{\rm{W}}}^v}{\bm{{\rm{W}}}^{vT}} = \bm{{\rm{I}}}
\end{aligned}
\end{gather}
where we also need to consider the condition $\bm{{\rm{BB}}}^T=\bm{{\rm{I}}}$ which is used during the optimization process. Specifically, we firstly convert Eq. \ref{15} to trace of matrix form, and then we utilize the SVD algorithm \cite{hu2018discrete} to solve the optimization problem.
\begin{equation}
\label{16}
{\bm{{\rm{W}}}^v} = \bm{{\rm{S}}}{\bm{{\rm{D}}}^T}
\end{equation}
where $\bm{{\rm{S}}}$ and $\bm{{\rm{D}}}$ are the left and right singular vectors of the compact Singular Value Decomposition (SVD) of $\bm{{\rm{Z}}}^v\bm{{\rm{B}}}^T$.

$\bm{{\rm{Updating}} }\quad \bm{{\rm{B}}}$: The common gradient method is not suitable for solving the discrete binary codes $\bm{{\rm{B}}}$. We rewrite Eq\ref{11} related to $\bm{{\rm{B}}}$ as follows:
\begin{gather}
\begin{aligned}
&\max_{\bm{{\rm{B}}}} tr({\bm{{\rm{B}}}^T}(2\sum\limits_{v=1}^M {{{(\bm{{\rm{p}}}^v)}^k}{\bm{{\rm{W}}}^v}{\bm{{\rm{Z}}}^v}} + \lambda \bm{{\rm{QH}}}))\\&
s.t. \quad\bm{{\rm{B}}} \in \left\{ { - 1,} \right.{\left. { + 1} \right\}^{b \times N}},\bm{{\rm{BB}}}^T = N\bm{{\rm{I}}}
\end{aligned}
\end{gather}

We need to preserve constraints of $\bm{B}$ which can generate compact and effective binary code. Thus, the optimal binary code $\bm{B}$ can be obtained as follows, and $sgn(\cdot)$ means symbolic function.
\begin{gather}
\label{18}
\bm{{\rm{B}}} =  sgn(\sum\limits_{v = 1}^M {(2{{({\bm{{\rm{p}}} ^v})}^k}} {\bm{{\rm{W}}}^v}{\bm{{\rm{Z}}}^v}) + \lambda \bm{{\rm{QH}}})
\end{gather}

$\bm{{\rm{Updating}} }\quad \bm{{\rm{Q}}}\quad and\quad \bm{{\rm{H}}}$: In this part, we iteratively optimize the binary clustering model by matrix factorization in Hamming space. By removing the irrelevant terms, the problem can be rewritten as:
\begin{gather}
\begin{aligned}
\label{19}
&\mathop {\min}\limits_{\bm{{\rm{Q,H}}}} \left\| {\bm{{\rm{B}}} - \bm{{\rm{QH}}} } \right\|_F^2 \\&s.t.\quad{\bm{{\rm{Q}}}^T}\bm{1} = \bm{0},\bm{{\rm{Q}}} \in \left\{ { - 1,} \right.{\left. { + 1} \right\}^{b \times c}},\bm{{\rm{H}}} \in \left\{ {0,} \right.{\left. 1 \right\}^{c \times N}},\\&\sum\limits_{i=1}^c {{\bm{{\rm{h}}}_{is}} = 1} ,\quad\bm{{\rm{B}}} \in \left\{ { - 1,} \right.{\left. { + 1} \right\}^{b \times N}}
\end{aligned}
\end{gather}

We simply reformulate Eq. \ref{19} to:
\begin{gather}
\begin{aligned}
&\mathop {\min}\limits_{\bm{{\rm{Q,H}}}} \left\|{\bm{{\rm{B}}} - \bm{{\rm{QH}}}} \right\|_F^2 + \rho {\left\| {{\bm{{\rm{Q}}}^T}\bm{1}} \right\|^2}\\&s.t.\quad \bm{{\rm{Q}}} \in \left\{ { - 1,} \right.{\left. { + 1} \right\}^{b \times c}},\bm{{\rm{H}}} \in \left\{ {0,} \right.{\left. 1 \right\}^{c \times N}},\sum\limits_{i=1}^c {{\bm{{\rm{h}}}_{is}} = 1} 
\end{aligned}
\end{gather}
which is equal Eq. \ref{19} with adaptively large $\rho$. The proposed matrix factorization strategy iteratively optimizes the cluster centroids and indicators as follows.

$\bm{{\rm{Updating}} }\quad \bm{{\rm{Q}}}\quad $: Due to the discrete constraint on $\bm{{\rm{Q}}}$, we utilize the discrete proximal linearized minimization (DPLM) \cite{shen2016fast} method, which can effectively obtain high-quality binary solutions. With $\bm{{\rm{H}}}$ fixed, the optimal $\bm{{\rm{Q}}}$ can be obtained as follow:
\begin{gather}
\begin{aligned}
{\min}{\cal L_{\bm{{\rm{Q}}}}} &= \left\| {\bm{{\rm{B}}} - \bm{{\rm{QH}}}} \right\|_F^2 + \rho {\left\| {{\bm{{\rm{Q}}}^T}\bm{1}} \right\|^2}\\ &
=  - 2tr({\bm{{\rm{B}}}^T}\bm{{\rm{QH}}}) + \rho {\left\| {{\bm{{\rm{Q}}}^T}} \right\|^2} + con\\ &s.t.\quad\bm{{\rm{Q}}} \in \left\{ { - 1,} \right.{\left. { + 1} \right\}^{l \times c}}
\end{aligned}
\end{gather}

According to the DPLM, in the $t+1$-th iteration $\bm{{\rm{Q}}}$ can be updated as:
\begin{gather}
\label{22}
{\bm{{\rm{Q}}}^{t + 1}} = {\mathop{ sgn}} ({\bm{{\rm{Q}}}^t} - \frac{1}{\mu }\nabla {\cal L}_{\bm{{\rm{Q}}}^t})
\end{gather}
where $\nabla {\cal L}_{\bm{{\rm{Q}}}^t}$ represents the gradient of $\cal L_{\bm{{\rm{Q}}}}$

$\bm{{\rm{Updating}} }\quad \bm{{\rm{H}}}\quad $: We utilize vectors-based method to optimize the indicator matrix $\bm{{\rm{H}}}$, the solution to $\bm{{\rm{h}}}_{i,j}$ can be easily obtained by
\begin{gather}
\label{23}
\bm{{\rm{h}}}_{i,j}^{t + 1} = \left\{ \begin{array}{l}
1,\quad j = \arg {\min _s}D({\bm{{\rm{b}}}_i},\bm{{\rm{q}}}_s^{t + 1})\\
0,\quad otherwise
\end{array} \right.
\end{gather}
where $D({\bm{{\rm{b}}}_i},\bm{{\rm{q}}}_s^{t + 1})$ is the distance between $i$-th binary code $\bm{{\rm{b}}}_i$ and the $s$-th cluster centroid $\bm{{\rm{q}}}_s$ in Hamming space. Notably, we use binary code rather than real-value to calculate Eq. \ref{23}, which adopt Hamming distance that can save time compared to the Euclidean distance.

$\bm{{\rm{Updating}} }\quad \bm{{\rm{p}}}^v$: For simplicity, we let ${{\rm{a}} ^v} = \left\| {{\bm{{\rm{W}}}^v}{\bm{{\rm{Z}}}^v}- \bm{{\rm{B}}}} \right\|_F^2+\| {{\bm{{\rm{Z}}}^v}-{\bm{{\rm{W}}}^{vT}}\bm{{\rm{B}}}} \|_F^2$. Based on the attributes of different views, the weighting coefficient $\bm{{\rm{p}}}^v$ can be equivalent as following equation:
\begin{gather}
\label{24}
\mathop {{\rm{min}}}\limits_{{\bm{{\rm{p}}}^v}} \sum\limits_{v = 1}^M {{{({\bm{{\rm{p}}}^v})}^k}} {{\rm{a}} ^v} \qquad s.t. \quad \sum\limits_{v=1}^M {\bm{{\rm{p}}} ^v = 1,} \bm{{\rm{p}}} ^v > 0
\end{gather}

Due to the constraint on $\bm{{\rm{p}}}^v$, we can solve this problem by the Lagrange multiplier method. By setting the Lagrange multiplier $\Gamma$, Eq. \ref{24} can be rewritten as:
\begin{gather}
\min {\cal L}({\bm{{\rm{p}}} ^v},\Gamma ) = \sum\limits_{v = 1}^M {{{({\bm{{\rm{p}}} ^v})}^k}} {{\rm{a}}^v} - \Gamma (\sum\limits_{v = 1}^M {{\bm{{\rm{p}}} ^v}}  - 1)
\end{gather}
and then, we calculate the partial derivative of ${\cal L}({\bm{p} ^v},\Gamma )$ about $\bm{p}^v$ and $\Gamma$, we can get:
\begin{gather}
\left\{ \begin{array}{l}
\frac{{\partial {\cal L}}}{{\partial {\bm{{\rm{p}}} ^v}}} \quad= k{(\bm{{\rm{p}}}^v)^{k - 1}}{\rm{a}}^v - \Gamma \\
\frac{{\partial {\cal L}}}{{\partial \Gamma }} \quad= \sum\limits_{v = 1}^M {(\bm{{\rm{p}}} ^{v})^k}  - 1
\end{array} \right.
\end{gather}

Therefore, we set partial derivative to zero which can get:
\begin{gather}
\label{27}
{\bm{{\rm{p}}} ^v} = \frac{{{{({\rm{a}}^v)}^{\frac{1}{{1 - k}}}}}}{{\sum\limits_{v=1}^M {{{({\rm{a}}^v)}^{\frac{1}{{1 - k}}}}} }}\end{gather}

We have presented the whole optimization process for Eq. \ref{11}. And we summarize the whole process in Algorithm 2, which iteratively updates the variables until convergence.
\begin{algorithm}[htbp!] 
	\caption{GCAE Algorithm}  
	\label{alg:B}  
	\begin{algorithmic}[1] 
		\REQUIRE {Dataset by kernelized $\phi({\bm{{\rm{X}}}^v});$ the parameter $\lambda$; the auxiliary matrices $\bm{{\rm{F}}}^v$ and $\bm{{\rm{G}}}^v$;} 
		\ENSURE  Binary code $\bm{{\rm{B}}}$; cluster indicator matrix $\bm{{\rm{H}}}$;  
		\STATE Randomly initialize binary code $\bm{{\rm{B}}}$; affinity graph $\bm{{\rm{Z}}}^v$; weights for different view $\bm{{\rm{p}}}^v$; binary code length ${\rm{b}}$; project matrix $\bm{{\rm{W}}}^v$
		\REPEAT 
		\STATE Update  $\bm{{\rm{Z}}}^v$ according to Eq. \ref{12}
		\STATE Update $\bm{{\rm{W}}}^v$ according to Eq. \ref{16}
		\STATE Update $\bm{{\rm{B}}}$ according to Eq. \ref{18}
		\STATE Update $\bm{{\rm{Q}}}$ and $\bm{{\rm{H}}}$ according to Eq. \ref{22} and Eq. \ref{23}
		\STATE Update $\bm{{\rm{p}}}^v$ according to Eq. \ref{27}
		\UNTIL convergance  
	\end{algorithmic}  
\end{algorithm}

\begin{table*}[tbp!]
	\centering
	\caption{The Comparison Result with Hash Method.} 
	\label{Table2}
	\resizebox{\textwidth}{6cm}{
		\begin{tabular}{cccccccccccc}
			\toprule
			Datasets & Metrics   & SH     & DSH    & SP     & ITQ    & SGH    & RSSH   & RFDH   & HSIC   & BMVC   & GCAE   \\ \midrule
			\multirow{6}{*}{100leaves}   & ACC & 0.4713 & 0.4494 & 0.4750 & 0.4569 & 0.5088 & 0.3631 & 0.4513 & 0.6563 & 0.4981 & \textbf{0.8888} \\
			& NMI       & 0.7214 & 0.7320 & 0.7240 & 0.7405 & 0.7579 & 0.6203 & 0.7161 & 0.8245 & 0.7291 & \textbf{0.9426} \\
			& Purity    & 0.4950 & 0.4988 & 0.5138 & 0.4944 & 0.5338 & 0.3888 & 0.4838 & 0.6788 & 0.5331 & \textbf{0.9031} \\
			& F-score   & 0.3471 & 0.3219 & 0.3312 & 0.3324 & 0.3996 & 0.2209 & 0.3234 & 0.5431 & 0.3057 & \textbf{0.8366} \\
			& Precision & 0.3184 & 0.2653 & 0.2942 & 0.2528 & 0.3626 & 0.2094 & 0.2708 & 0.5128 & 0.2547 & \textbf{0.8166} \\
			& ARI       & 0.3404 & 0.3141 & 0.3240 & 0.3240 & 0.3933 & 0.2131 & 0.3158 & 0.5385 & 0.2978 & \textbf{0.8350} \\ \midrule
			\multirow{6}{*}{Caltech-101} & ACC & 0.1747 & 0.1610 & 0.2077 & 0.2472 & 0.2256 & 0.2860 & 0.2196 & 0.2429 & 0.2930 & \textbf{0.3005} \\
			& NMI       & 0.3252 & 0.3600 & 0.4034 & 0.4404 & 0.4349 & 0.4846 & 0.4424 & 0.4451 & \textbf{0.4900} & 0.4711 \\
			& Purity    & 0.3089 & 0.3439 & 0.3878 & 0.4231 & 0.4051 & 0.4729 & 0.4228 & 0.4125 & \textbf{0.4907} & 0.4421 \\
			& F-score   & 0.1486 & 0.1250 & 0.1738 & 0.2311 & 0.2089 & 0.2533 & 0.2081 & 0.2055 & 0.2465 & \textbf{0.3023} \\
			& Precision & 0.2604 & 0.1975 & 0.2897 & 0.3544 & 0.3085 & 0.4071 & 0.3419 & 0.3430 & 0.4147 & \textbf{0.4229} \\
			& ARI       & 0.1347 & 0.1091 & 0.1596 & 0.2167 & 0.1935 & 0.2406 & 0.1943 & 0.1919 & 0.2336 & \textbf{0.2880} \\ \midrule
			\multirow{6}{*}{Cifar-10}    & ACC & 0.1708 & 0.2189 & 0.2275 & 0.2245 & 0.2205 & 0.1960 & 0.2252 & 0.2153 & 0.2350 & \textbf{0.2498} \\
			& NMI       & 0.0282 & 0.0938 & 0.0979 & 0.0987 & 0.0995 & 0.0659 & 0.1011 & 0.0920 & 0.1016 & \textbf{0.1020} \\
			& Purity    & 0.1727 & 0.2271 & 0.2285 & 0.2297 & 0.2218 & 0.2046 & 0.2333 & 0.2236 & 0.2368 & \textbf{0.2549} \\
			& F-score   & 0.1137 & 0.1462 & 0.1342 & 0.1266 & 0.1458 & 0.1316 & 0.1560 & 0.1441 & 0.1590 & \textbf{0.1597} \\
			& Precision & 0.1129 & 0.1520 & 0.1495 & 0.1544 & 0.1408 & 0.1284 & 0.1457 & 0.1420 & 0.1533 & \textbf{0.1563} \\
			& ARI       & 0.0145 & 0.0624 & 0.0584 & 0.0610 & 0.0600 & 0.0325 & 0.0582 & 0.0476 & 0.0618 & \textbf{0.0642} \\ \midrule
			\multirow{6}{*}{SUNRGBD}     & ACC & 0.1164 & 0.1577 & 0.1925 & 0.1895 & 0.1823 & 0.1613 & 0.1738 & 0.1616 & 0.1379 & \textbf{0.2423} \\
			& NMI       & 0.1319 & 0.2198 & 0.2180 & 0.2108 & 0.2172 & 0.1980 & 0.2032 & 0.2202 & 0.1545 & \textbf{0.2207} \\
			& Purity    & 0.2441 & 0.3271 & 0.3418 & 0.3366 & 0.3433 & 0.3279 & 0.3332 & \textbf{0.3524} & 0.2803 & 0.3435 \\
			& F-score   & 0.0650 & 0.1033 & 0.1223 & 0.1274 & 0.1184 & 0.0976 & 0.1145 & 0.1059 & 0.0822 & \textbf{0.1541} \\
			& Precision & 0.1193 & 0.1861 & 0.1722 & 0.1802 & 0.1822 & 0.1869 & 0.1901 & \textbf{0.2013} & 0.1550 & 0.1909 \\
			& ARI       & 0.0309 & 0.0699 & 0.0895 & 0.0951 & 0.0859 & 0.0661 & 0.0823 & 0.0744 & 0.0496 & \textbf{0.1076} \\ \midrule
			\multirow{6}{*}{Caltech256}  & ACC & 0.0622 & 0.0782 & 0.0856 & 0.0865 & 0.0826 & 0.0937 & 0.0783 & 0.0678 & 0.0932 & \textbf{0.1071} \\
			& NMI       & 0.2557 & 0.2866 & 0.2947 & 0.2717 & 0.2969 & 0.3058 & 0.2855 & 0.2350 & \textbf{0.3185} & 0.2871 \\
			& Purity    & 0.1096 & 0.1255 & 0.1399 & 0.1386 & 0.1363 & 0.1512 & 0.1278 & 0.1064 & \textbf{0.1534} & 0.1415 \\
			& F-score   & 0.0465 & 0.0644 & 0.0635 & 0.0975 & 0.0623 & 0.0805 & 0.0610 & 0.0402 & 0.0745 & \textbf{0.1145} \\
			& Precision & 0.0527 & 0.0681 & 0.0658 & 0.0950 & 0.0649 & 0.0932 & 0.0644 & 0.0305 & 0.0811 & \textbf{0.1037} \\
			& ARI       & 0.0415 & 0.0591 & 0.0582 & 0.0920 & 0.0569 & 0.0758 & 0.0558 & 0.0327 & 0.0695 & \textbf{0.1087} \\ \bottomrule
	\end{tabular}}
\end{table*}

\section{Experiments}
In this section, we describe the datasets and comparison methods to verify the effectiveness of the proposed GCAE for clustering tasks. We evaluated the performance of GCAE by comparing several hash methods and multi-view clustering methods on $5$ widely used datasets. Moreover, we analyze the parameter sensitivity of our model, which can affect the fluctuations of results. We also summarize the running times on all datasets and evaluate the convergence of our model. All the experiments are conducted using Matlab 2020a on a Windows PC with Intel 2.8-GHz CPU and 64-GB RAM.
\subsection{Experimental settings}
In this part, we describe the utilized datasets and the comparison methods in detail. We also introduce some evaluation metrics which aim to verify the effectiveness of GCAE.
\subsubsection{Datasets}
To evaluate the clustering performance of the proposed model and comparison models, five widely-accepted multi-view datasets are selected. The details of the selected datasets are as follows:

\textbf{Caltech101\footnote{http://www.vision.caltech.edu/ImageDatasets/Caltech101/}} contains 9144 samples of 101 objects. 6 publicly available features are utilized as multiple views, including Gabor feature with 48 dimension, LBP feature with 928 dimension, GIST feature with 512 dimension, CENTRIST feature with 254 dimension, wavelet moments(WM) with 40 dimension and HOG feature with 1984 dimension.

\textbf{Caltech256\footnote{http://www.vision.caltech.edu/Image-Datasets/Caltech256}} contains 30607 images by 4 kinds of features. It includes 256 classes with more than 80 samples per class. Besides, each image is represented by color histogram with 729 dimension, Gist feature with 1024 dimension, HOG feature with 1152 dimension and features based on convolution network with 1440 dimension, which are four different types of presentation. 

\textbf{Cifar-10\footnote{https://www.cs.toronto.edu/kriz/cifar.html.}} is composed by 60000 tiny color images in 10 kinds of classes. Specifically, we represent the features by DSD feature with 220 dimension, HOG feature with 512 dimension and Gist feature with 768 dimension. Besides, a subset of this dataset is selected in the experiment that contains 10000 samples.

\textbf{100leaves\footnote{https://archive.ics.uci.edu/ml/dataset} \cite{mallah2013plant}} contains 1600 samples from each of 100 plant species, which form the UCI repository with three different views, i.e., Shape, Texture and Margin features.

\textbf{SUNRGBD\footnote{http://rgbd.cs.princeton.edu/}} contains 10335 indoor scene images by 3D cameras in 45 classes. Similar with \cite{wang2021fast}, we utilize two views with 4096 dimension and features extracted by different convolution network.
\begin{table*}[htbp!]
	\centering
	\caption{The Clustering Results on Five Datasets.} 
	\label{Table3}
	\resizebox{\textwidth}{5.5cm}{
		\begin{tabular}{cccccccccccccc}
			\toprule
			Datasets & Metrics   & k-means & SC     & Co-re-p & Co-re-c & AMGL   & Mul-NMF & MLAN   & mPAC    & GMC    & HSIC            & BMVC            & GCAE            \\ \midrule
			\multirow{6}{*}{100leaves}   & ACC & 0.6200 & 0.4894 & 0.7253 & 0.7939 & 0.7631 & 0.8694 & 0.7356 & 0.8238 & 0.4338 & 0.6563 & 0.4981 & \textbf{0.8888} \\
			& NMI       & 0.8284  & 0.7649 & 0.8835   & 0.9257   & 0.9065 & 0.9288  & 0.8848 & 0.9292  & 0.7014 & 0.8245          & 0.7291          & \textbf{0.9426} \\
			& Purity    & 0.6550  & 0.5575 & 0.7565   & 0.8272   & 0.8063 & 0.8981  & 0.7625 & 0.8506  & 0.5350 & 0.6788          & 0.5331          & \textbf{0.9031} \\
			& F-score   & 0.5233  & 0.2144 & 0.6595   & 0.7558   & 0.4513 & 0.8236  & 0.6583 & 0.5042  & 0.3145 & 0.5431          & 0.3057          & \textbf{0.8366} \\
			& Precision & 0.4689  & 0.1307 & 0.6107   & 0.7011   & 0.3086 & 0.7832  & 0.6082 & 0.3521  & 0.3744 & 0.5128          & 0.2547          & \textbf{0.8166} \\
			& ARI       & 0.5183  & 0.2021 & 0.6560   & 0.7533   & 0.4437 & 0.8219  & 0.6548 & 0.4974  & 0.3070 & 0.5385          & 0.2978          & \textbf{0.8350} \\ \midrule
			\multirow{6}{*}{Caltech-101} & ACC & 0.1331 & 0.1751 & 0.2611 & 0.2587 & 0.1476 & 0.1908 & 0.2274 & 0.1950 & 0.2672 & 0.2429 & 0.2940 & \textbf{0.3005} \\
			& NMI       & 0.3078  & 0.3207 & 0.4752   & 0.4912   & 0.3757 & 0.3519  & 0.4564 & 0.3446  & 0.4408 & 0.4451          & \textbf{0.4900} & 0.4711          \\
			& Purity    & 0.2907  & 0.3107 & 0.4622   & 0.4664   & 0.1681 & 0.3184  & 0.4401 & 0.3012  & 0.3497 & 0.4125          & \textbf{0.4907} & 0.4421          \\
			& F-score   & 0.0985  & 0.1362 & 0.2202   & 0.2226   & 0.0338 & 0.0470  & 0.1930 & 0.0496  & 0.2658 & 0.2055          & 0.2465          & \textbf{0.3023} \\
			& Precision & 0.1351  & 0.1440 & 0.3954   & 0.3999   & 0.0175 & 0.0248  & 0.3185 & 0.0261  & 0.2337 & 0.3430          & 0.4147          & \textbf{0.4229} \\
			& ARI       & 0.0796  & 0.1126 & 0.2078   & 0.2103   & 0.0155 & -0.0068 & 0.1790 & -0.0042 & 0.2475 & 0.1919          & 0.2336          & \textbf{0.2880} \\ \midrule
			\multirow{6}{*}{Cifar-10}    & ACC & 0.2036 & 0.1712 & 0.2169 & 0.2084 & 0.2232 & 0.1217 & 0.2304 & 0.1038 & 0.2312 & 0.2153 & 0.2350 & \textbf{0.2498} \\
			& NMI       & 0.0891  & 0.0773 & 0.0944   & 0.0903   & 0.0845 & 0.0204  & 0.0919 & 0.0066  & 0.1014 & 0.0920          & 0.1016          & \textbf{0.1020} \\
			& Purity    & 0.2055  & 0.1751 & 0.2214   & 0.2180   & 0.2276 & 0.1235  & 0.2522 & 0.1043  & 0.2462 & 0.2236          & 0.2368          & \textbf{0.2549} \\
			& F-score   & 0.1529  & 0.1443 & 0.1593   & 0.1477   & 0.1587 & 0.1582  & 0.1570 & 0.1511  & 0.1499 & 0.1441          & 0.1590          & \textbf{0.1597} \\
			& Precision & 0.1344  & 0.1080 & 0.1453   & 0.1393   & 0.1393 & 0.1005  & 0.1537 & 0.0999  & 0.1495 & 0.1420          & 0.1533          & \textbf{0.1563} \\
			& ARI       & 0.0443  & 0.0155 & 0.0561   & 0.0467   & 0.0594 & 0.0012  & 0.0612 & 0.0603  & 0.0607 & 0.0476          & 0.0618          & \textbf{0.0642} \\ \midrule
			\multirow{6}{*}{SUNRGBD}     & ACC & 0.1859 & 0.1060 & 0.1824 & 0.1829 & 0.1010 & 0.1346 & 0.1898 & 0.1277 & 0.2155 & 0.1616 & 0.1379 & \textbf{0.2423} \\
			& NMI       & 0.1866  & 0.0084 & 0.2109   & 0.2161   & 0.1883 & 0.0976  & 0.2101 & 0.0728  & 0.2204 & 0.2202          & 0.1545          & \textbf{0.2207} \\
			& Purity    & 0.3022  & 0.1087 & 0.3274   & 0.3360   & 0.1119 & 0.1583  & 0.3257 & 0.1415  & 0.2303 & \textbf{0.3524} & 0.2803          & 0.3435          \\
			& F-score   & 0.1293  & 0.1213 & 0.1172   & 0.1221   & 0.0637 & 0.1200  & 0.1209 & 0.1215  & 0.1387 & 0.1059          & 0.0822          & \textbf{0.1541} \\
			& Precision & 0.1057  & 0.0646 & 0.1849   & 0.1748   & 0.0364 & 0.0645  & 0.1822 & 0.0650  & 0.1007 & \textbf{0.2013} & 0.1550          & 0.1909          \\
			& ARI       & 0.0976  & 0.0323 & 0.0865   & 0.0916   & 0.0262 & 0.0251  & 0.0891 & 0.0008  & 0.1026 & 0.0744          & 0.0496          & \textbf{0.1076} \\ \midrule
			\multirow{6}{*}{Caltech-256} & ACC & 0.0845 & 0.0924 & 0.0854 & 0.1025 & 0.0467 & 0.0612 & 0.0761 & 0.0904   & 0.0723   & 0.0678 & 0.0932 & \textbf{0.1071} \\
			& NMI       & 0.2748  & 0.2764 & 0.2997   & 0.2786   & 0.1070 & 0.1486  & 0.2820 & 0.2198     & 0.1347   & 0.2350          & \textbf{0.3185} & 0.2871          \\
			& Purity    & 0.1296  & 0.1339 & 0.1412   & 0.1522   & 0.0415 & 0.1060  & 0.1343 & 0.1381    & 0.1008  & 0.1064          & \textbf{0.1534} & 0.1415          \\
			& F-score   & 0.0562  & 0.0628 & 0.0596   & 0.0713   & 0.0466 & 0.0106  & 0.0556 & 0.0498      & 0.0355   & 0.0402          & 0.0745          & \textbf{0.1145} \\
			& Precision & 0.0522  & 0.0855 & 0.0692   & 0.0813   & 0.0458 & 0.0053  & 0.0570 & 0.0412      & 0.0515   & 0.0305          & 0.0811          & \textbf{0.1037} \\
			& ARI       & 0.0501  & 0.0846 & 0.0549   & 0.0689   & 0.0488 & -0.0011 & 0.0501 & 0.0361      & 0.0834    & 0.0327          & 0.0695          & \textbf{0.1087} \\ \bottomrule
	\end{tabular}}
\end{table*}

\subsubsection{Comparing Algorithms and Evaluation Metrics}
In our experiments, we evaluate the performance of GCAE by comparing several state-of-the-art multi-view algorithms and hash algorithms. Specifically, we utilize seven single-view hash methods and two multi-view hash clustering algorithms, including SH \cite{weiss2008spectral}, DSH \cite{jin2013density}, SP \cite{xia2015sparse}, ITQ \cite{gong2012iterative}, SGH \cite{jiang2015scalable}, RSSH \cite{tian2020unsupervised}, RFDH \cite{wang2017robust}, HSIC \cite{zhang2018highly}, BMVC \cite{zhang2018binary}. For single-view hash methods, we utilize K-means algorithm to process the generated binary code for clustering and we preserve the best result of each view clustering. Besides, we adopt eight real-value multi-view clustering algorithms as comparing algorithms, including K-means \cite{hartigan1979algorithm}, SC \cite{ng2001spectral}, Co-regularize \cite{kumar2011co}, AMGL \cite{nie2016parameter}, Mul-NMF \cite{liu2013multi}, MLAN \cite{nie2017multi}, mPAC \cite{kang2019multiple}, GMC \cite{wang2019gmc}. We utilize the source code from the author's public homepage for comparing. Notably, we set the length of binary code as 128-bits in our experiments, which can effectively preserve critical information. 

In order to generally evaluate the superiority of our method, we utilize six widely used evaluation metrics, i.e., clustering accuracy(ACC), Normalized Mutual Information(NMI), Purity, F-score, Precision and ARI \cite{jiang2021graph}, \cite{shi2021multi}. For all algorithms, better performance is improved by the higher value of evaluation metrics.
\subsection{Experimental Results and Analysis}
In this section, we conducted experiments on $5$ multi-view datasets to show the superiority of the proposed GCAE. The detailed clustering results are shown in Table \uppercase\expandafter{\romannumeral2}, \uppercase\expandafter{\romannumeral3}. We utilize the bold values to represent the best performance in each table corresponding to the dataset. Besides, this section also introduced the analysis of parameter sensitivity, which can reflect clustering results with different parameters. Finally, we provide the complexity analysis and convergence analysis of the proposed model to verify the stability of GCAE.

\subsubsection{Comparison with Hash Methods}
In this section, we conduct experiments with hash methods on five multi-view datasets to verify the performance of our proposed model. We compare GCAE with seven single-view hash methods and two multi-view hash clustering methods. In single-view hash methods, we generate cluster results by adopting k-means algorithm to finish cluster task. Specifically, we adopt K-means algorithm with cluster binary code to obtain sample labels. The results with different datasets are reported in Table \ref{Table2}.

\begin{figure*}[tbp!]
	\centering
	\setlength{\belowcaptionskip}{-1mm}
	\vspace{-0.35cm} %设置与上面正文的距离
	\subfigtopskip=-1pt %设置子图与上面正文或别的内容的距离
	\subfigbottomskip=-1pt %设置第二行子图与第一行子图的距离，即下面的头与上面的脚的距离
	\subfigcapskip=-5pt %设置子图与子标题之间的距离
	\subfigure[Clustering accuracy with $\lambda$]{
		\includegraphics[width=4.3cm]{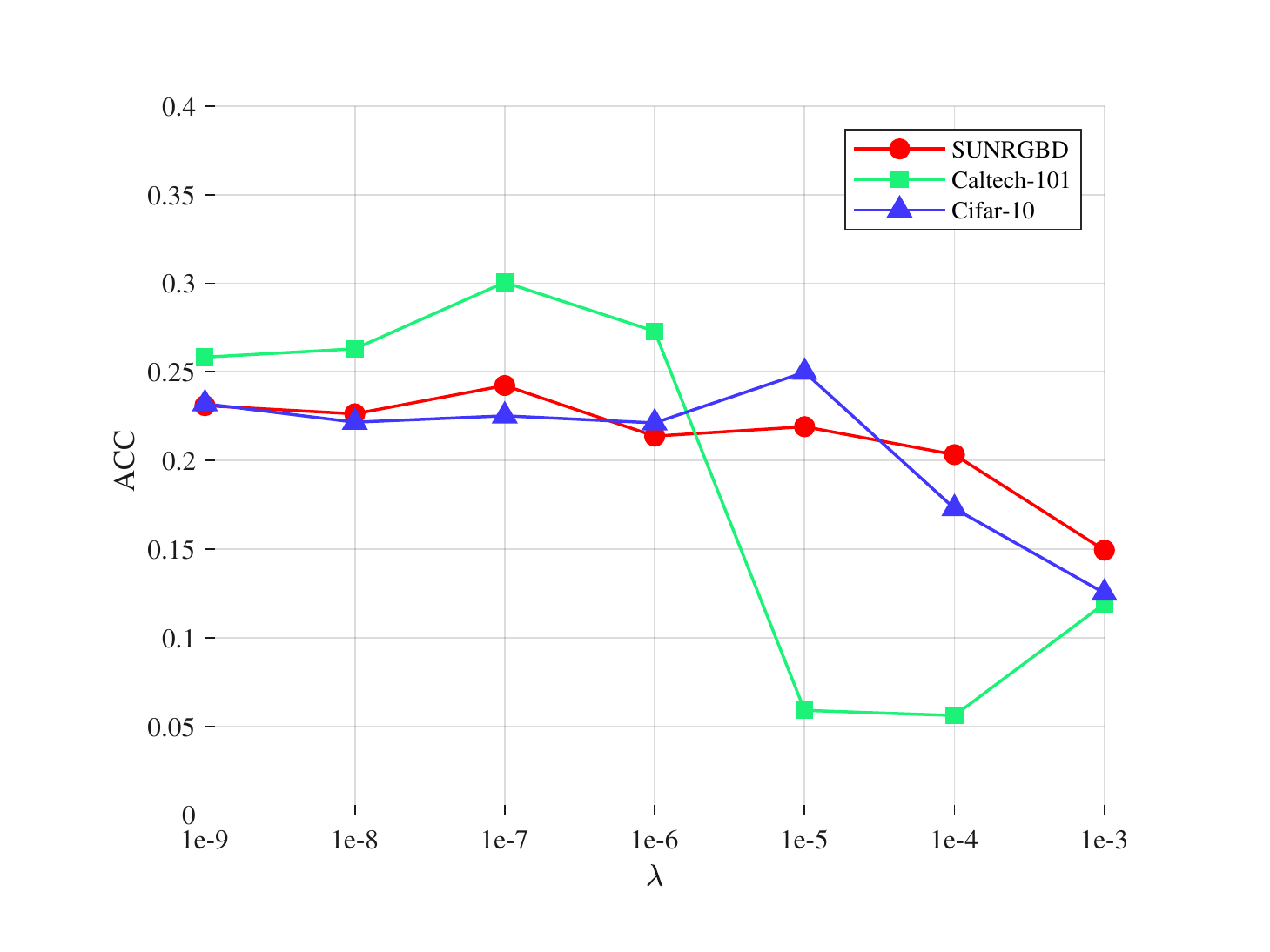}}
	\subfigure[Clustering accuracy with $k$]{
		\includegraphics[width=4.3cm]{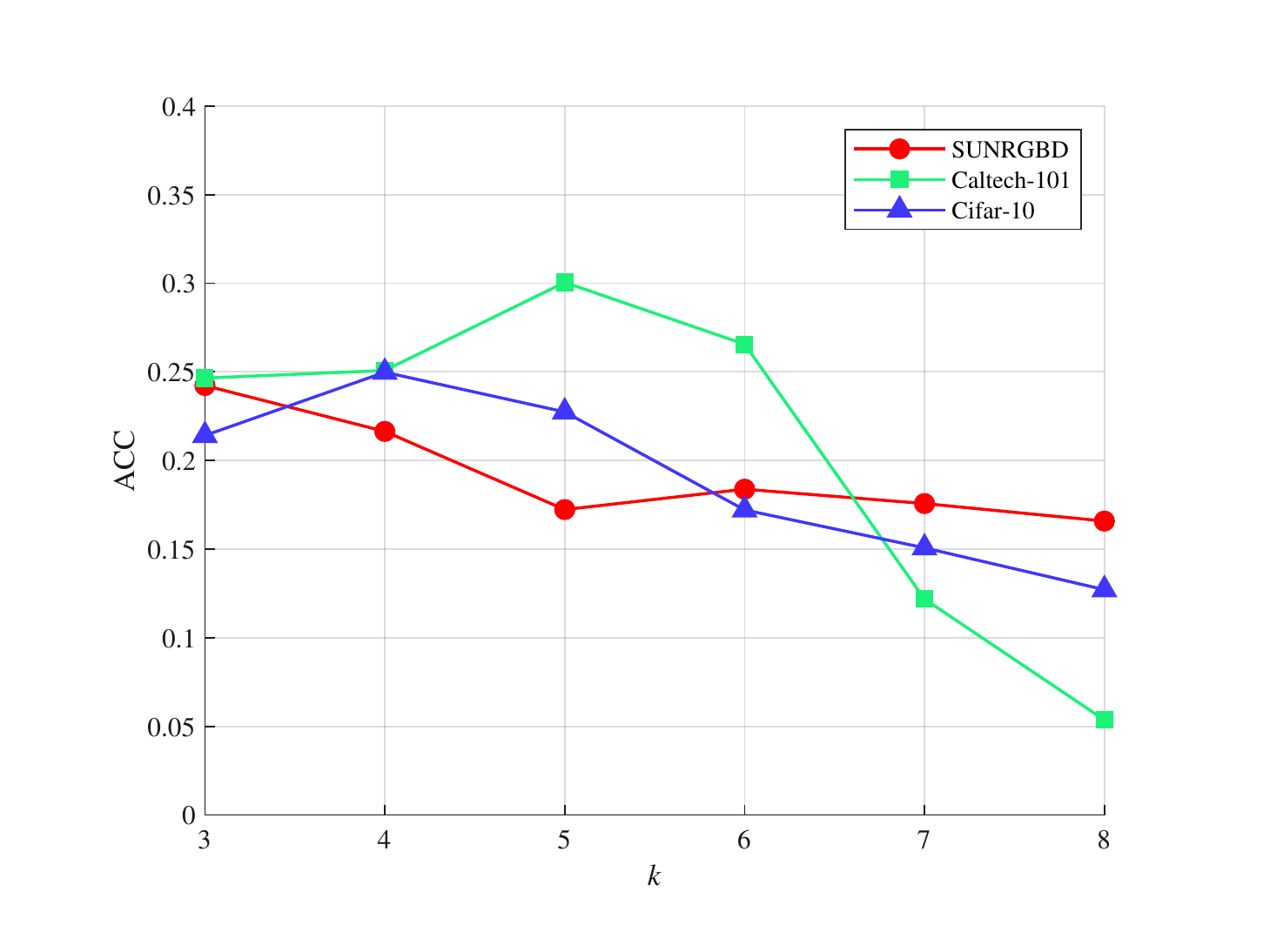}}
	\subfigure[Clustering accuracy with  $t$]{
		\includegraphics[width=4.3cm]{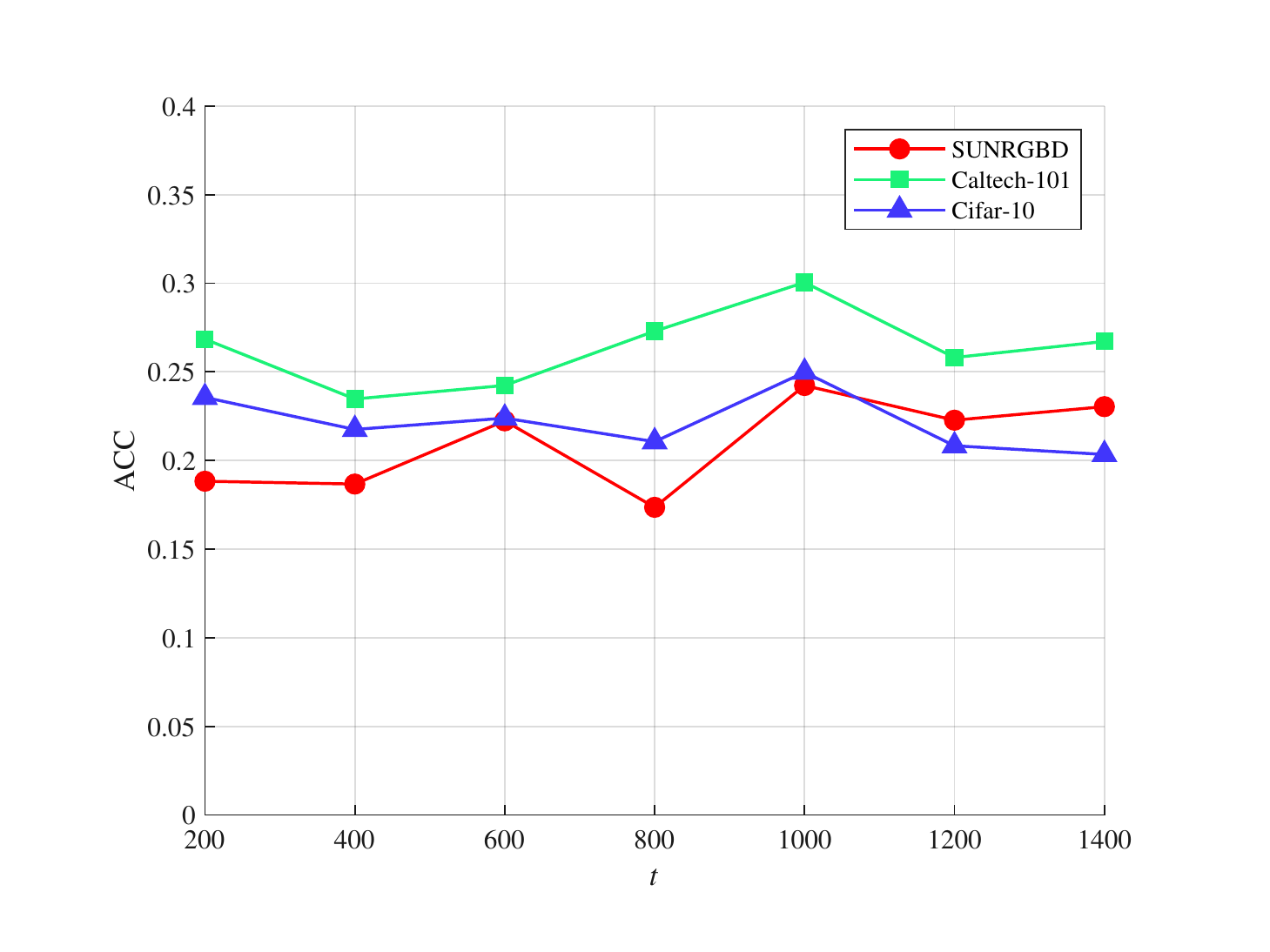}}
	\subfigure[Clustering accuracy with $\theta$]{
		\includegraphics[width=4.3cm]{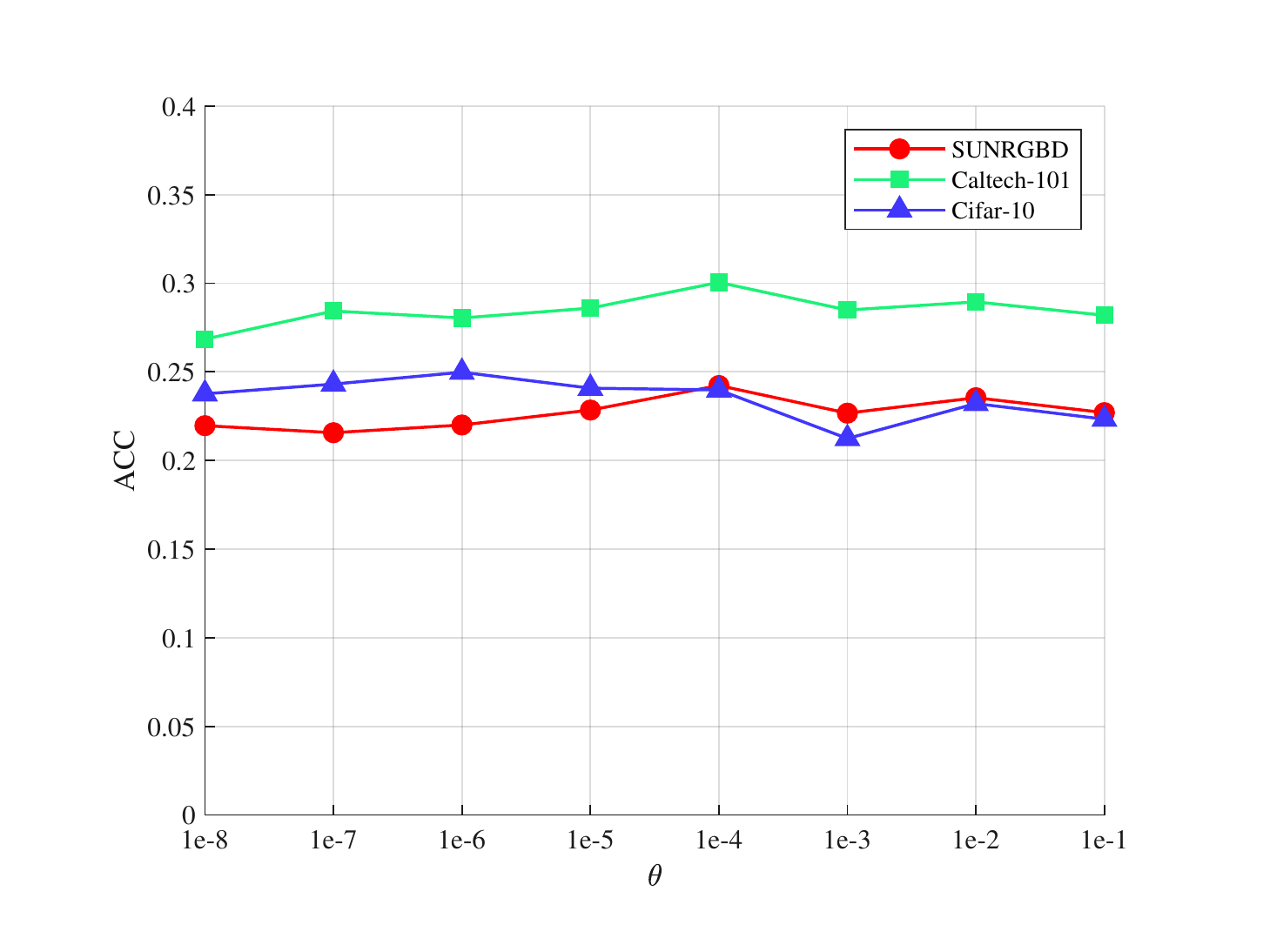}}			
	\caption{Clustering accuracy with parameter tuning for the Caltech-101, Cifar-10 and SUNRGBD datasets (a)$\lambda$ (b)$k$ (c)$t$ (d)$\theta$.   }
	\label{parameter1}
\end{figure*}

We summarized the clustering performance with compared methods on all multimedia datasets in Table \ref{Table2}, and the best results are highlighted in bold. As shown in Table \ref{Table2}, we can obvious that GCAE obtains the best clustering accuracy on five multi-view datasets compared with hash methods. Specifically, we proposed GCAE method evidently outperforms other state-of-the-art methods for 100leaves and Cifar-10 datasets. For the 100leaves dataset, the performance of GCAE improves around $23.2\%$, $11.8\%$, and $22.4\%$ with terms of ACC, NMI, and Purity over the second-best method. Moreover, for the Cifar-10 dataset, the performance improvements over the second-best method are $1.5\%$, $0.4\%$, $1.81\%$ with terms of ACC, NMI, and Purity. 

The experiments demonstrate that the multi-view hash clustering methods get better results than single-view methods, which proves multi-view methods can exploit complementary information from multi-view. For single-view methods, we perform clustering on each views and select the best results to report in the above table. Generally speaking, multi-view methods take the complementary information and consistent information between multi-view data into consideration. Therefore, multi-view methods can get better performance and single-view methods can not obtain satisfactory performance in most situations. The results in Table \ref{Table2} also verify the low-rank affinity graphs construction is a vital part of our proposed GCAE, which aims to effectively generate binary code with essential information.

Overall, the clustering performance about our proposed model outperforms the other hash methods in most situations. The phenomenon denotes that GCAE can effectively preserve essential information by auto-encoders structure and keep the discrete constraint on binary code. Besides, GCAE has the ability to integrate comprehensive information from multi-view data.

\subsubsection{Multi-view Methods Experimental Results and Analysis}
In this section, we present the detailed comparison clustering results with multi-view clustering methods in Table \ref{Table3}. As shown in this table, we compare GCAE with eight real-value clustering methods and two hash clustering methods. Moreover, for single-view methods i.e., K-means and SC, we concatenates all multiple views into one view for clustering.

As shown in Table \ref{Table3}, it is clear that GCAE obtains the best clustering accuracy with the comparison results on all multi-view datasets. Specifically, the proposed GCAE model outperforms all other methods on ACC, NMI, Purity, F-score, Precision, and ARI in most situations. We can obviously find the performance improvements over the second-best method are $1.9\%$, $1.4\%$ and $14.1\%$ respectively in metrics of ACC, NMI, and Purity for 100leaves dataset. Moreover, for the Cifar-10 dataset, our proposed model can improve $1.5\%$, $0.4\%$, and $1.81\%$ in the above metrics which is compared with the second-best method.

We conducted the experiments with comparing real-value based multi-view clustering method and hash methods, which aims to verify the stability of clustering in Hamming space. It is clear that hash methods can obtain satisfactory performance in most situations. This is because real-value methods utilize Euclidean distance to measure two samples, which have problems of low efficiency and high computational complexity. And hash methods learn binary code and obtain cluster results in Hamming space, which can improve calculation efficiency. The adoption of the affinity graph is more conductive to preserving vital information of mining original data, but utilizing hash methods only improves significantly in computational complexity, which can not significantly improve evaluation metrics.

\begin{figure*}[htbp!]
	\centering
	\setlength{\belowcaptionskip}{-1mm}
	\vspace{-0.35cm} %设置与上面正文的距离
	\subfigtopskip=-1pt %设置子图与上面正文或别的内容的距离
	\subfigbottomskip=-1pt %设置第二行子图与第一行子图的距离，即下面的头与上面的脚的距离
	\subfigcapskip=-5pt %设置子图与子标题之间的距离
	\subfigure[100leaves]{
		\includegraphics[width=5.5cm]{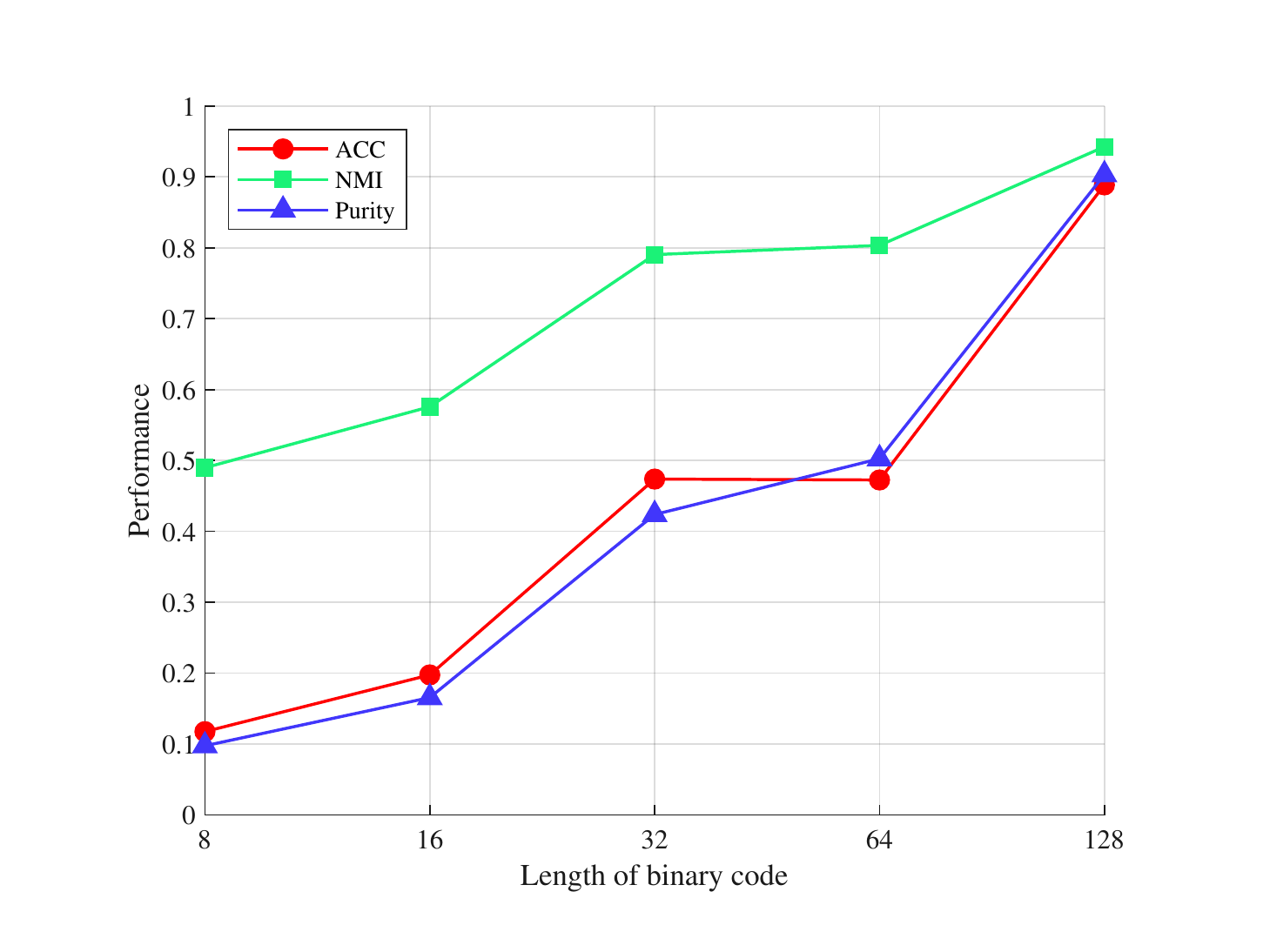}
		
	}
	\subfigure[Caltech-101]{
		\includegraphics[width=5.5cm]{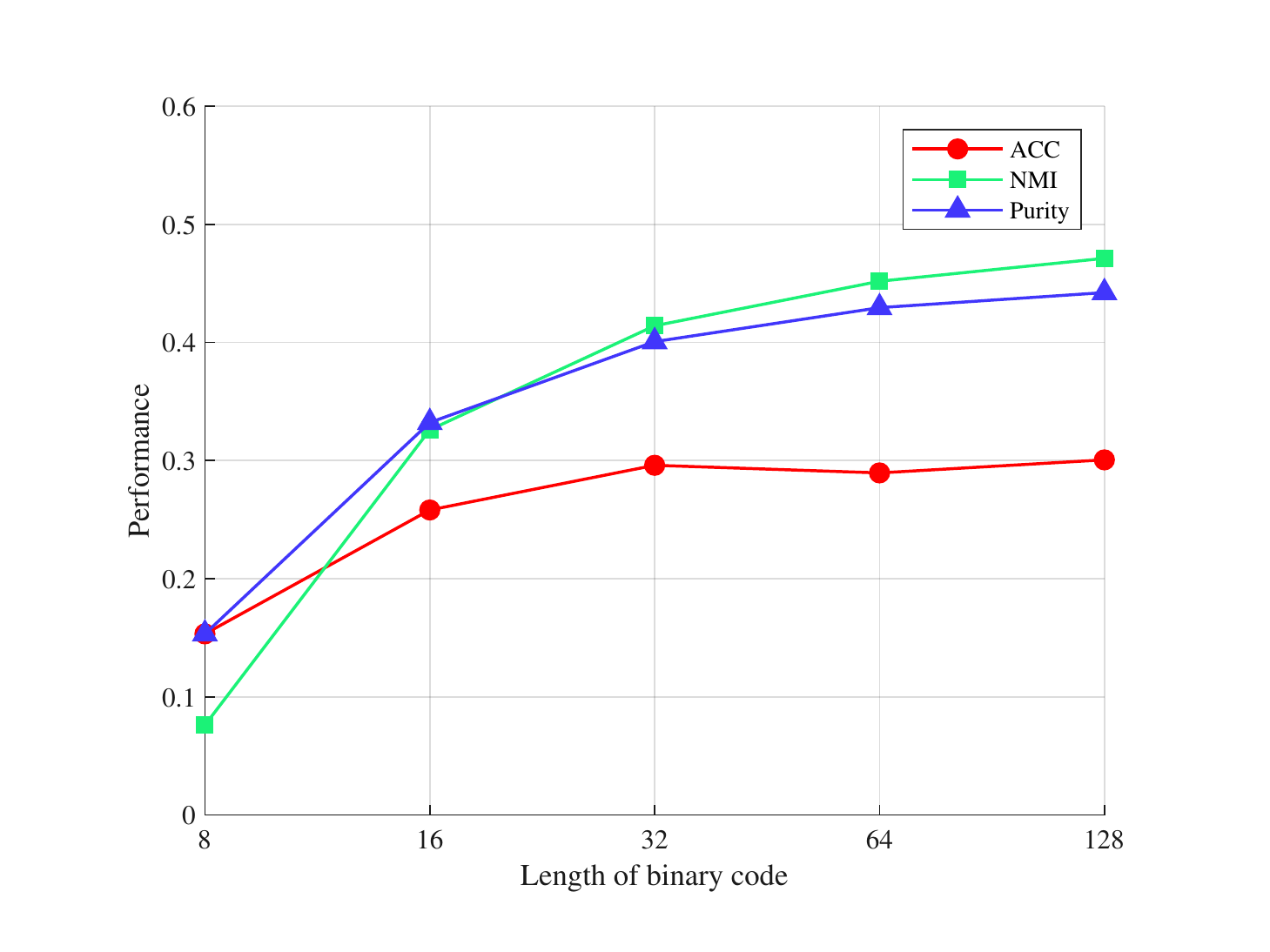}}
	\subfigure[Cifar-10]{
		\includegraphics[width=5.5cm]{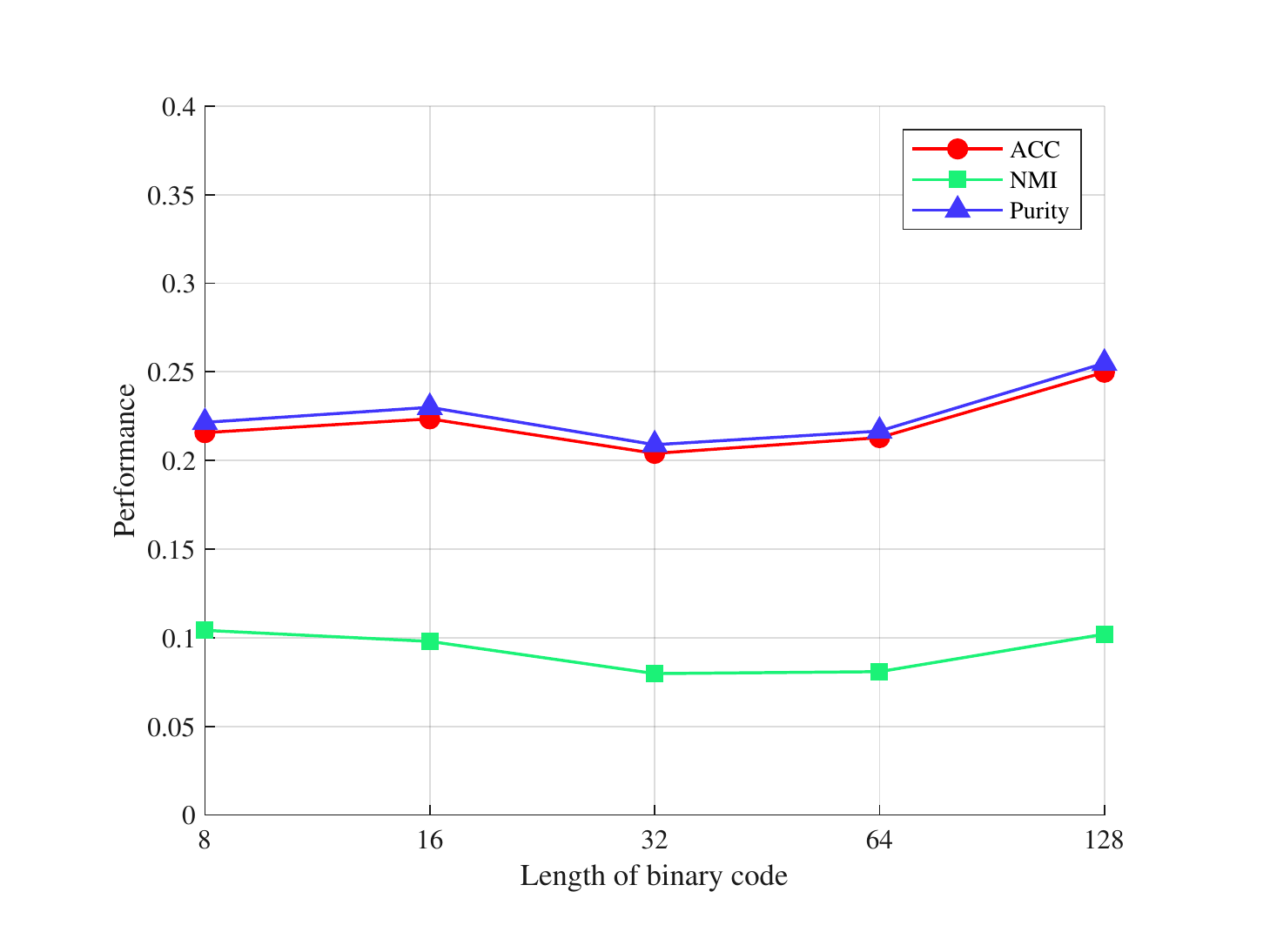}}
	
	\caption{From left to right, the performance with binary code length tuning for the 100leaves, Caltech-101 and Cifar-10 datasets, respectively.  }
	\label{binary}
\end{figure*}

It is worth noting that the proposed GCAE obtain the best cluster accuracy and satisfactory performance on other evaluation metrics for five multi-view datasets. This is because GCAE can explore consistency and complementary information in multi-view data structure by learning low-rank affinity graphs, which learn high-quality collaborative representation well and eliminates some redundant and noise information in the original real-valued features. Besides, GCAE can also improve the computational complexity of algorithm significantly. 

\begin{figure*}[htb]
	\centering
	\setlength{\belowcaptionskip}{-1mm}
	\vspace{-0.35cm} %设置与上面正文的距离
	\subfigtopskip=-1pt %设置子图与上面正文或别的内容的距离
	\subfigbottomskip=-1pt %设置第二行子图与第一行子图的距离，即下面的头与上面的脚的距离
	\subfigcapskip=-5pt %设置子图与子标题之间的距离
	\subfigure[100leaves]{
		\includegraphics[width=5.5cm]{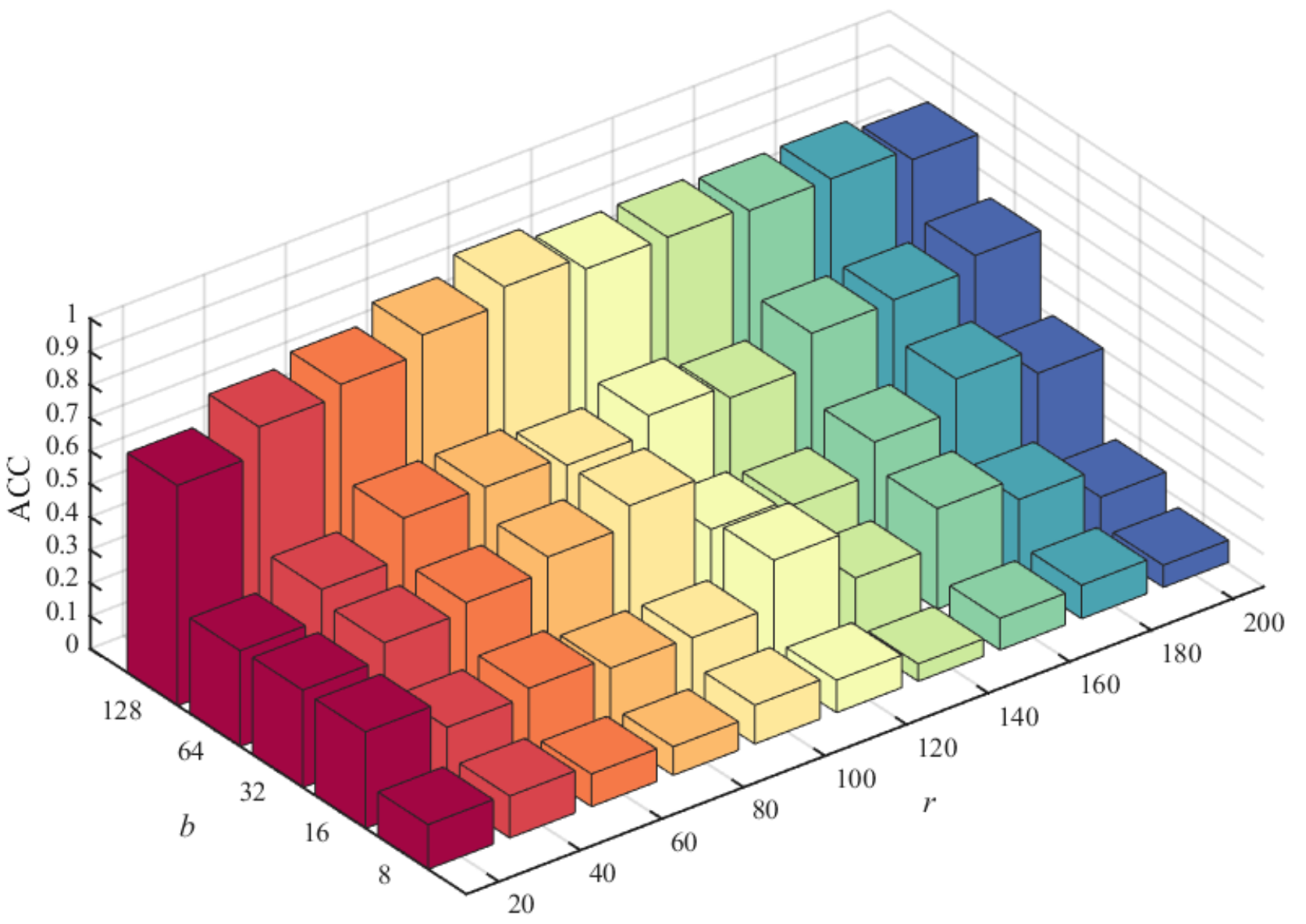}}
	\subfigure[Caltech-101]{
		\includegraphics[width=5.5cm]{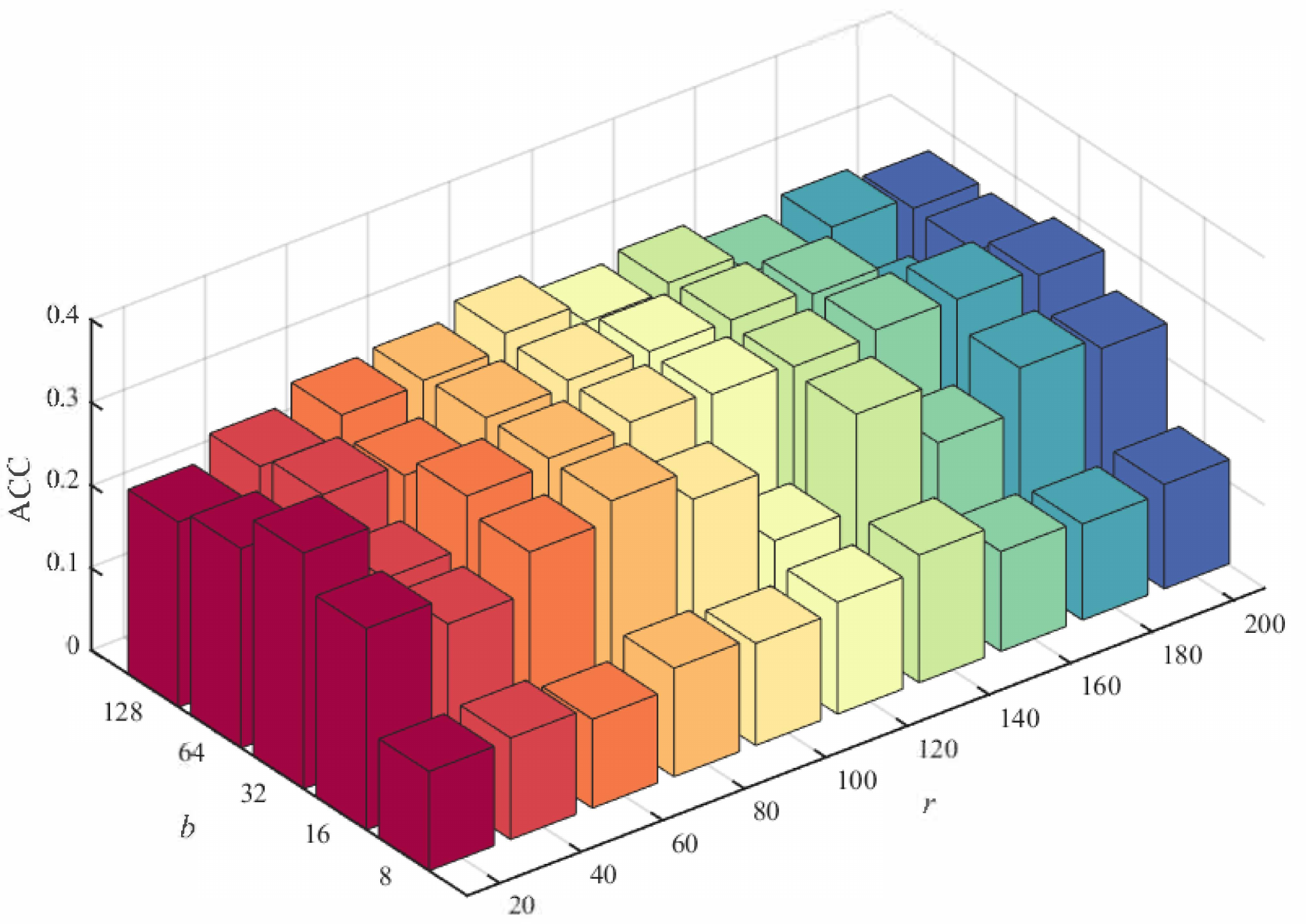}}
	\subfigure[Cifar-10]{
		\includegraphics[width=5.5cm]{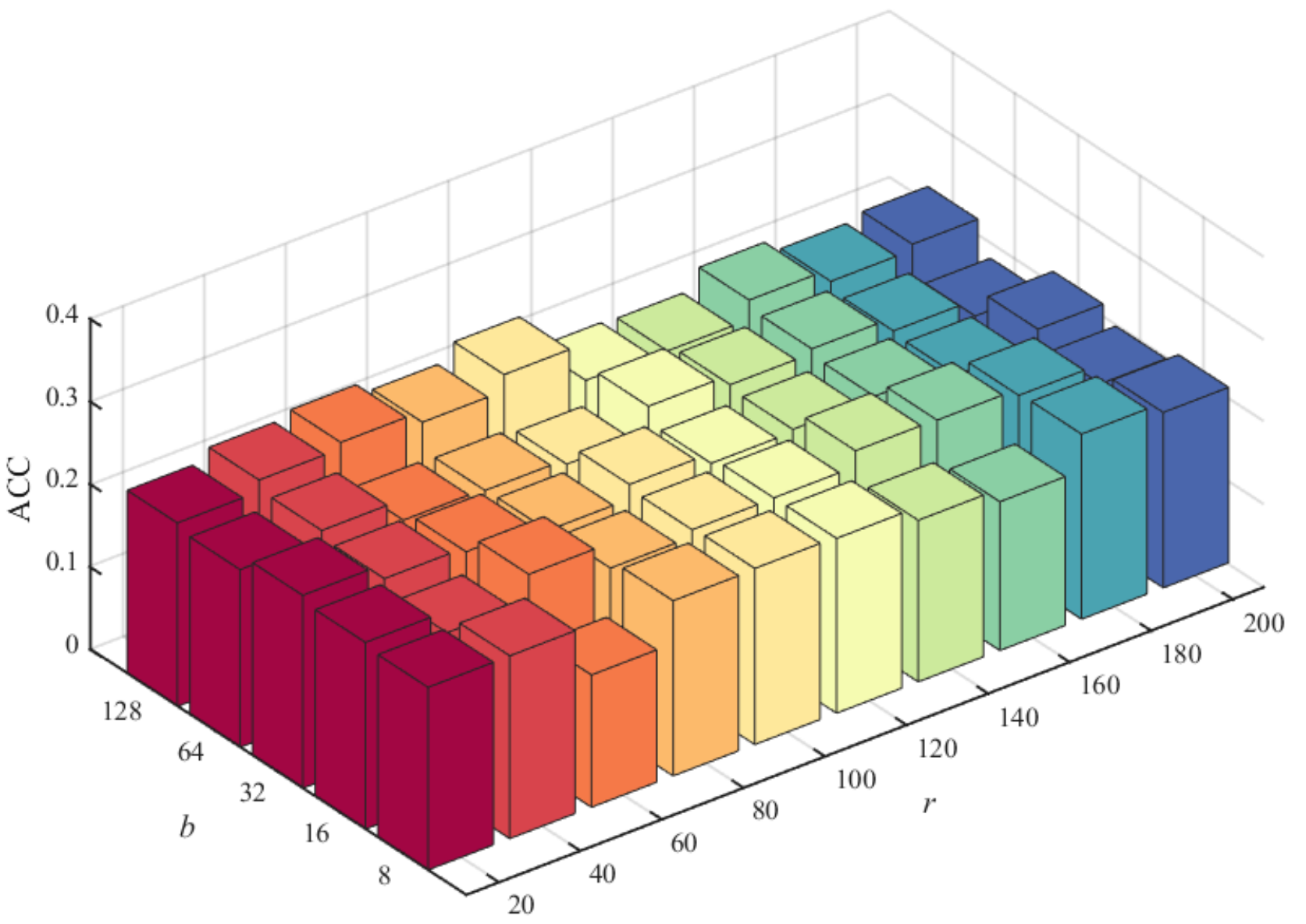}} 
	\caption{Parameter tuning with respect to $r$ and $b$ for the 100leaves, Caltech-101 and Cifar-10 datasets.}
	\label{3d}
\end{figure*}

\subsection{Parameter Sensitivity and Complexity Analysis}
In this section, we discover that the fluctuation of several parameters has a significant influence on the experiments. In order to evaluate the sensitivity of the parameters in the proposed GCAE, we select 100leaves, Caltech-101, Cifar-10, and SUNRGBD as the benchmark datasets. For GCAE, there are six parameters to be tuned, i.e., $\lambda,k,t,\theta,b$ and $r$. The above parameters represent regularization parameter for binary clustering, the power of normalized weighting coefficient, the dimension of nonlinear kernel, the small smooth item, binary bits, respectively. We firstly conducted experiments on the Caltech-101, Cifar-10, and SUNRGBD datasets to show the influence of parameters $\lambda,k,t$ and $\theta$ change trend on the results. Specifically, we calculate the cluster accuracy values and display them in line charts, which can intuitively find the influence of parameters changing. Fig. \ref{parameter1} shows the different change trends
on three benchmark datasets. In Fig. \ref{parameter1}(a), we can observe the different $\lambda$ values have effected the cluster accuracy notably on three datasets. We conduct experiments with different parameters $\lambda$ selected from the range of $[1e-9,1e-3]$. It is obvious that the change trend of GCAE on the three datasets and GCAE can achieve ideal performance when $\lambda$ is set to an appropriate value. Besides, it is obvious that the parameter $k$ can effect cluster accuracy values selected from the range of $[3,8]$ in Fig. \ref{parameter1}(b). The proposed GCAE can achieve excellent performance with a good selection of $k$.  As shown in Fig. \ref{parameter1}(c) and Fig. \ref{parameter1}(d), the change trend of $t$ and $\theta $ do not influence cluster results obviously. It is notable that GCAE can still achieve a relatively stable and good performance with the change in the dimension of nonlinear kernel $t$ and the small smooth item $\theta$, which are selected in range of $[200,1400]$ and $[1e^{-8},1e^{-1}]$, respectively.

As shown in Fig. \ref{binary}, we adopt comparison experiments of the length of binary code to verify our proposed GCAE. Therefore, we change the binary code length and calculate ACC, NMI, and Purity on 100leaves, Caltech-101 and Cifar-10 datasets to show the performance of GCAE. Above the figure, we select 128-bits binary code in GCAE, which aims to fully preserve essential information for clustering. Besides, in order to demonstrate the relation between the rank of affinity graphs and the length of binary code. We conducted a three-dimensional statistical graph of clustering results by adjusting the values of $r$ and $b$. The statistical results of 100leaves, Caltech-101 and Cifar-10 datasets are shown in Fig. \ref{3d}, which clustering performance as ACC. We can obviously find GCAE can obtain excellent performance with a good selection of $r$ and $b$, which are designed in the range of $[20,200]$ and $[8,128]$ respectively. In conclusion, we select 128-bits binary codes and $r=100$ on these datasets, which achieves the best clustering performance with the proposed GCAE algorithm.

We summarized the overall algorithm of the proposed GCAE method in Algorithm 1 and Algorithm 2, which present low-rank affinity graphs learning and graph-collaborated auto-encoder hashing respectively. The computational burden of GCAE mainly consists of the above two sub-process, and we provide the analysis of GCAE in detail. Specifically, the two closed-form solutions, i.e., Eq. \ref{9} and Eq. \ref{10} cost $O(Nd^{v}r+Nr^2+r^3)$. Therefore, the whole computational complexity of Algorithm 1 is $O(N\times max(d^v,r)\times r\times Iter)$, where we set the $Iter$ to 80. For Algorithm 2, we adopt the iterative optimization method to obtain the solution for each parameter. When updating $\bm{Z}^v$, which is the most time-consuming part of GCAE. It costs $O(N^3+N^2b+N^2r)$ because it needs to calculate the inverse of the matrix. And then, we need $O(N^2b)$ for updating $\bm{W}^v$, which contains SVD operation and matrix multiplication. After that, in order to update $\bm{B}$ and \ref{23}, we need $O(N^2b+Nbc)$ which includes Frobenius norm and ADPLM method. Besides, solving Eq. \ref{22} and Eq. \ref{23} needs $O(Nc)$ and updating $\bm{l}^v$ costs $O(rN)$ for each iteration. Overall, the whole computational complexity of GCAE is $O((N^3+3N^2b+N^2r+Nbc)p+Nmax(d^v,r)rIter)$, which $p$ is the number of epoch. For clarity, we summarized the running time of GCAE for each dataset in Table \ref{Table4}.
\begin{table}[htbp!]
	\centering
	\caption{Running time of GCAE.}
	\label{Table4}
	\scalebox{0.8}{
		\begin{tabular}{cccccc}
			\toprule
			Datasets & 100leaves & Caltech-101 & Cifar-10 & SUNRGBD & Caltech-256 \\ \midrule
			Times(s) & 14.16     & 1158.71     & 1120.15  & 1488.33 & 47618       \\ \bottomrule
	\end{tabular}}
\end{table}
\subsection{Convergence Analysis}
\begin{figure}[tbp!]
	\centering
	\setlength{\belowcaptionskip}{-1mm}
	\vspace{-0.35cm} %设置与上面正文的距离
	\subfigtopskip=-1pt %设置子图与上面正文或别的内容的距离
%	\subfigbottomskip=-1pt %设置第二行子图与第一行子图的距离，即下面的头与上面的脚的距离
	\subfigcapskip=-5pt %设置子图与子标题之间的距离
	\subfigure[100leaves]{
		\includegraphics[width=4.2cm]{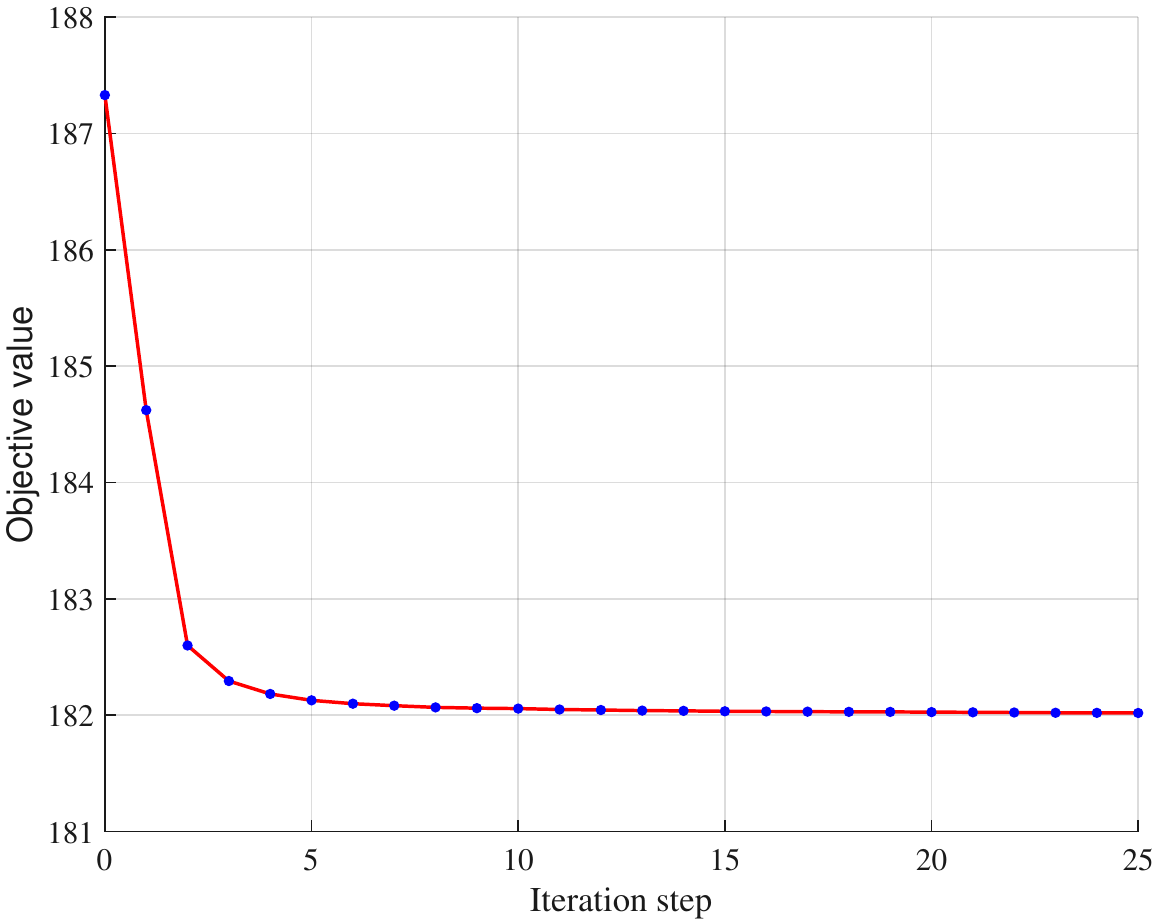}}
	\subfigure[Caltech-101]{
		\includegraphics[width=4.2cm]{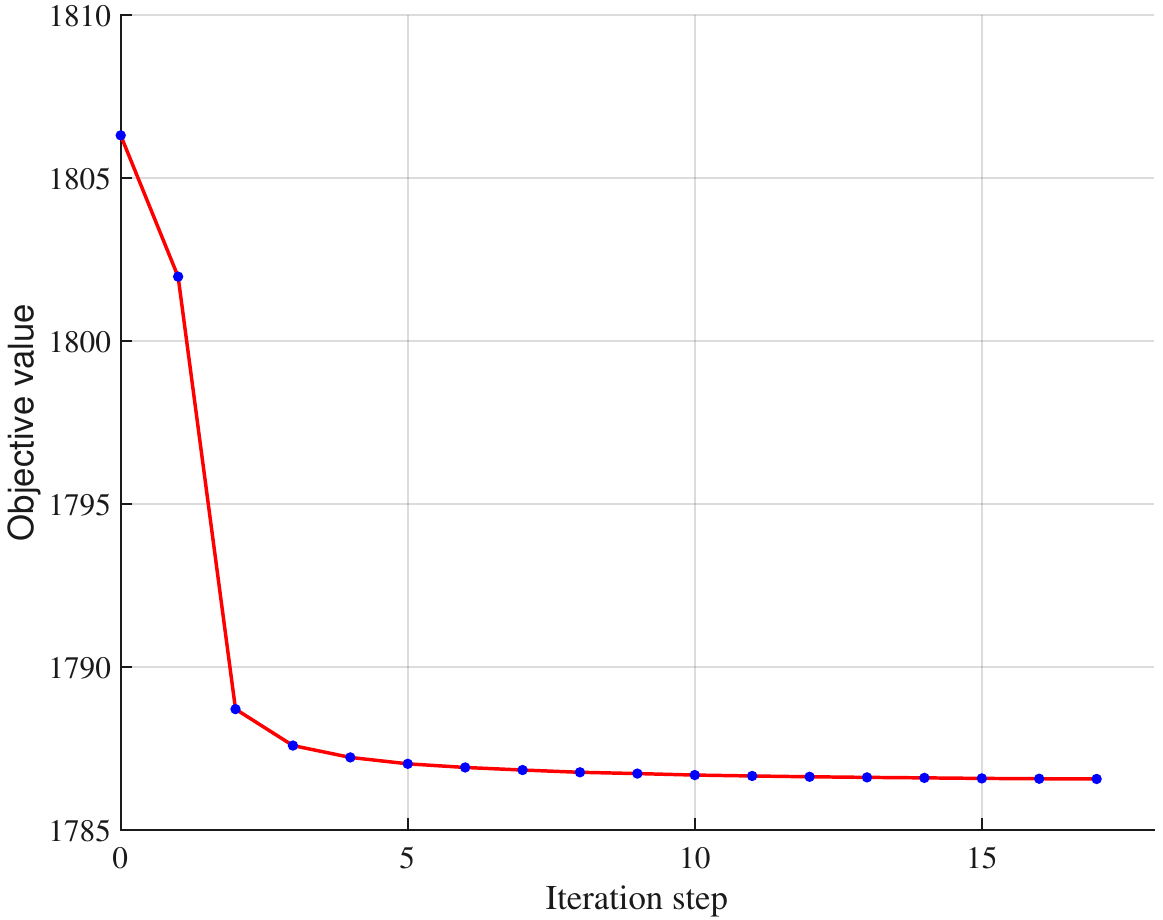}}
	\subfigure[Cifar-10]{
		\includegraphics[width=4.2cm]{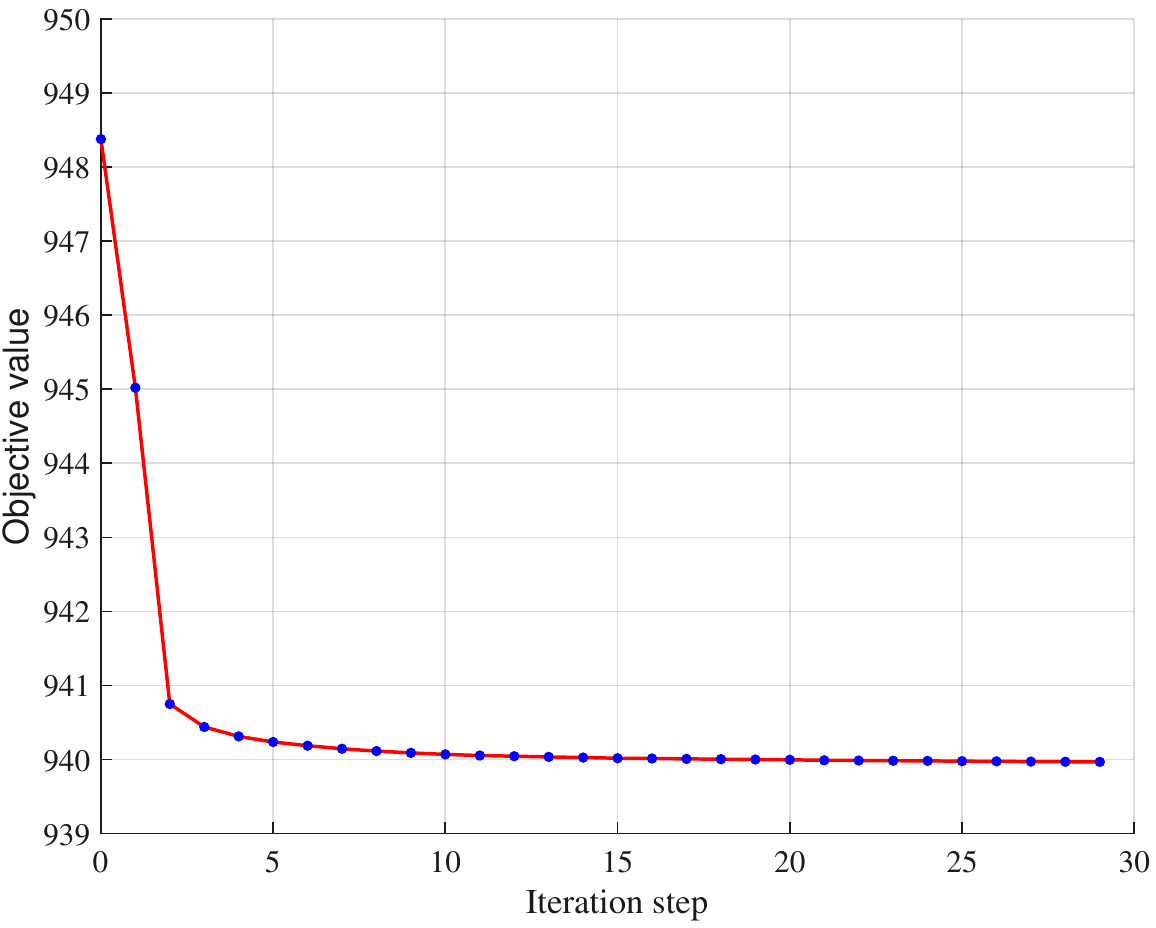}}
	\subfigure[SUNRGBD]{
		\includegraphics[width=4.2cm]{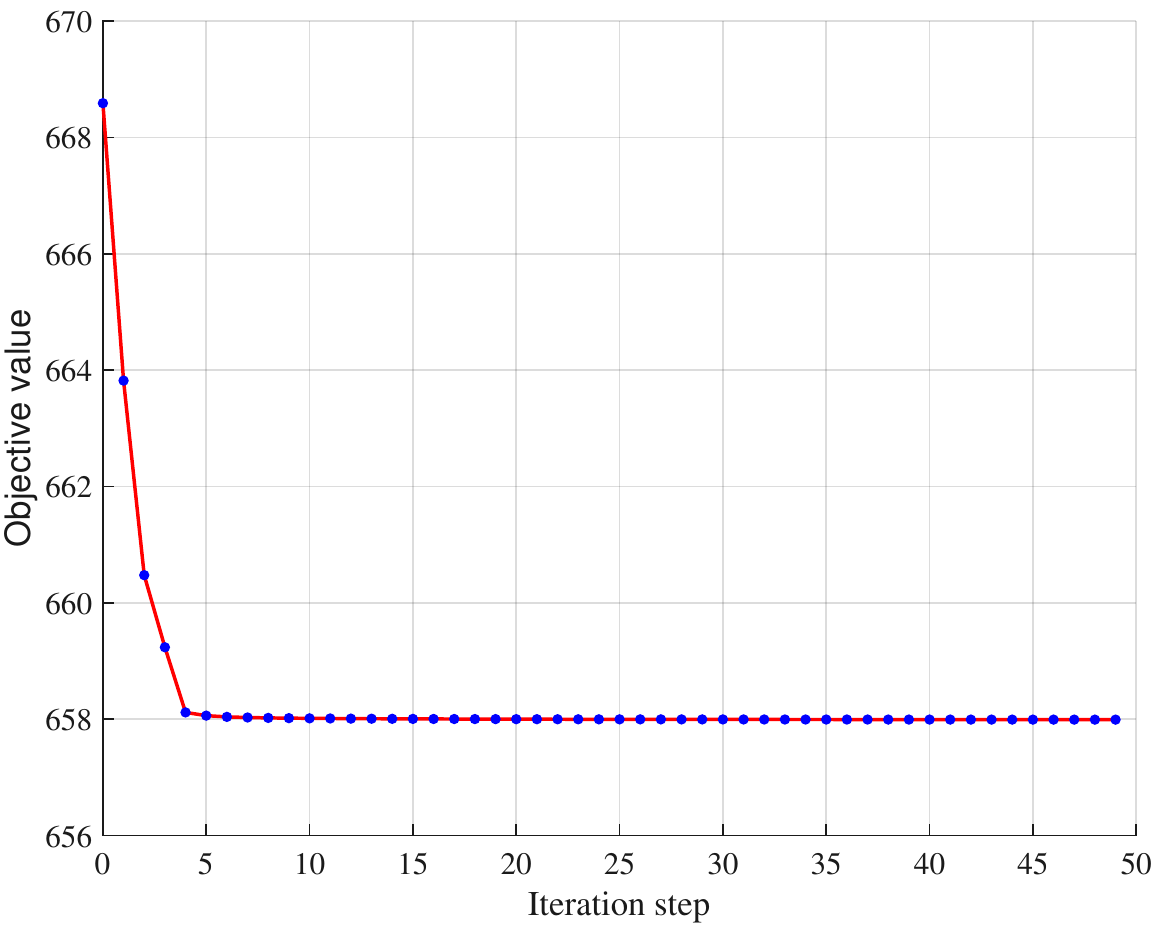}}			
	\caption{Convergence results for the 100leaves, Caltech-101, Cifar-10 and SUNRGBD datasets, respectively.  }
	\label{loss}
\end{figure}

In order to verify the proposed iteratively optimization algorithm for GCAE is convergent, we conduct the convergence analysis of the proposed GCAE in this section. Fig. \ref{loss} shows the convergence curves of GCAE on the datasets 100leaves, Caltech-101 and Cifar-10. In Fig .\ref{loss}, the $x$-axis and $y$-axis present the number of iterations and the value of the objective function, respectively. It can obviously find that the values of the objective function in GCAE decrease monotonically with the increased iterations. Furthermore, the objective values converge very quickly after 5-10 iterations which verifies the effectiveness of our proposed optimized solution for each sub-process.
\section{Conclusion}
In this paper, we proposed a novel binary multi-view clustering method termed as Graph-Collaborated Auto-encoder Hashing for Multi-view Binary Clustering (GCAE), which can effectively construct low-rank affinity graphs from each view and jointly learn binary code by auto-encoders. GCAE constructs graphs by utilizing auxiliary matrices to control the low-rank constraint, which can reasonably preserve essential information from original data. And then GCAE adopts auto-encoders to collaborate multiple graphs for learning unified binary codes and obtaining cluster results. With the proposed optimization algorithm, the objective function converges quickly. The extensive experiments demonstrated the superiority of GCAE. It can be obviously seen that different from real-value multi-view clustering methods, GCAE can effectively obtain cluster results in Hamming space. Compared with hash methods, GCAE can reasonably utilize multi-view consistency and complementarity information. We conducted our experiments on five multi-view datasets for experimental examination. In the future work, we intend to further improve the speed of the algorithm, which is the deficiency of the proposed method.

\section*{Acknowledgment}

This work was supported in part by the National Natural Science Foundation of China Grant 62002041 and 62176037, Liaoning Fundamental Research Funds for Universities Grant LJKQZ2021010, Liaoning Doctoral Research Startup Fund Project Grant 2021-BS-075,  Liaoning Province Applied Basic Research Project 22JH2/101300264 and Dalian Science and Technology Innovation Fund 2022JJ12GX019, 2021JJ12GX028 and 2022JJ12GX016.

%\IEEEpubidadjcol

\bibliographystyle{IEEEtran}
\bibliography{mybibfile}

\begin{thebibliography}{10}
\providecommand{\url}[1]{#1}
\csname url@samestyle\endcsname
\providecommand{\newblock}{\relax}
\providecommand{\bibinfo}[2]{#2}
\providecommand{\BIBentrySTDinterwordspacing}{\spaceskip=0pt\relax}
\providecommand{\BIBentryALTinterwordstretchfactor}{4}
\providecommand{\BIBentryALTinterwordspacing}{\spaceskip=\fontdimen2\font plus
\BIBentryALTinterwordstretchfactor\fontdimen3\font minus
  \fontdimen4\font\relax}
\providecommand{\BIBforeignlanguage}[2]{{%
\expandafter\ifx\csname l@#1\endcsname\relax
\typeout{** WARNING: IEEEtran.bst: No hyphenation pattern has been}%
\typeout{** loaded for the language `#1'. Using the pattern for}%
\typeout{** the default language instead.}%
\else
\language=\csname l@#1\endcsname
\fi
#2}}
\providecommand{\BIBdecl}{\relax}
\BIBdecl

\bibitem{zhang2022autoencoder}
C.~Zhang, Y.~Geng, Z.~Han, Y.~Liu, H.~Fu, and Q.~Hu, ``Autoencoder in
  autoencoder networks,'' \emph{IEEE transactions on neural networks and
  learning systems}, 2022.

\bibitem{zhang2021dual}
Y.~Zhang, Z.~Zhang, Y.~Wang, Z.~Zhang, L.~Zhang, S.~Yan, and M.~Wang,
  ``Dual-constrained deep semi-supervised coupled factorization network with
  enriched prior,'' \emph{International Journal of Computer Vision}, vol. 129,
  no.~12, pp. 3233--3254, 2021.

\bibitem{wang2018multiview}
Y.~Wang, L.~Wu, X.~Lin, and J.~Gao, ``Multiview spectral clustering via
  structured low-rank matrix factorization,'' \emph{IEEE transactions on neural
  networks and learning systems}, vol.~29, no.~10, pp. 4833--4843, 2018.

\bibitem{qian2022switchable}
B.~Qian, Y.~Wang, H.~Yin, R.~Hong, and M.~Wang, ``Switchable online knowledge
  distillation,'' in \emph{European Conference on Computer Vision}.\hskip 1em
  plus 0.5em minus 0.4em\relax Springer, 2022, pp. 449--466.

\bibitem{wang2021survey}
Y.~Wang, ``Survey on deep multi-modal data analytics: Collaboration, rivalry,
  and fusion,'' \emph{ACM Transactions on Multimedia Computing, Communications,
  and Applications (TOMM)}, vol.~17, no.~1s, pp. 1--25, 2021.

\bibitem{peng2020deep}
X.~Peng, J.~Feng, J.~T. Zhou, Y.~Lei, and S.~Yan, ``Deep subspace clustering,''
  \emph{IEEE transactions on neural networks and learning systems}, vol.~31,
  no.~12, pp. 5509--5521, 2020.

\bibitem{wang2015robust}
Y.~Wang, X.~Lin, L.~Wu, W.~Zhang, Q.~Zhang, and X.~Huang, ``Robust subspace
  clustering for multi-view data by exploiting correlation consensus,''
  \emph{IEEE Transactions on Image Processing}, vol.~24, no.~11, pp.
  3939--3949, 2015.

\bibitem{guo2010completed}
Z.~Guo, L.~Zhang, and D.~Zhang, ``A completed modeling of local binary pattern
  operator for texture classification,'' \emph{IEEE transactions on image
  processing}, vol.~19, no.~6, pp. 1657--1663, 2010.

\bibitem{movellan2002tutorial}
J.~R. Movellan, ``Tutorial on gabor filters,'' \emph{Open Source Document},
  vol.~40, pp. 1--23, 2002.

\bibitem{dalal2005histograms}
N.~Dalal and B.~Triggs, ``Histograms of oriented gradients for human
  detection,'' in \emph{2005 IEEE computer society conference on computer
  vision and pattern recognition (CVPR'05)}, vol.~1.\hskip 1em plus 0.5em minus
  0.4em\relax Ieee, 2005, pp. 886--893.

\bibitem{lowe2004distinctive}
D.~G. Lowe, ``Distinctive image features from scale-invariant keypoints,''
  \emph{International journal of computer vision}, vol.~60, no.~2, pp. 91--110,
  2004.

\bibitem{gu2019clustering}
Y.~Gu, S.~Wang, H.~Zhang, Y.~Yao, W.~Yang, and L.~Liu, ``Clustering-driven
  unsupervised deep hashing for image retrieval,'' \emph{Neurocomputing}, vol.
  368, pp. 114--123, 2019.

\bibitem{wang2022progressive}
Y.~Wang, J.~Peng, H.~Wang, and M.~Wang, ``Progressive learning with multi-scale
  attention network for cross-domain vehicle re-identification,'' \emph{Science
  China Information Sciences}, vol.~65, no.~6, pp. 1--15, 2022.

\bibitem{hu2018deep}
W.~Hu, Y.~Fan, J.~Xing, L.~Sun, Z.~Cai, and S.~Maybank, ``Deep constrained
  siamese hash coding network and load-balanced locality-sensitive hashing for
  near duplicate image detection,'' \emph{IEEE Transactions on Image
  Processing}, vol.~27, no.~9, pp. 4452--4464, 2018.

\bibitem{wang2022towards}
H.~Wang, G.~Jiang, J.~Peng, R.~Deng, and X.~Fu, ``Towards adaptive consensus
  graph: Multi-view clustering via graph collaboration,'' \emph{IEEE
  Transactions on Multimedia}, 2022.

\bibitem{jiang2022tensorial}
G.~Jiang, J.~Peng, H.~Wang, Z.~Mi, and X.~Fu, ``Tensorial multi-view clustering
  via low-rank constrained high-order graph learning,'' \emph{IEEE Transactions
  on Circuits and Systems for Video Technology}, 2022.

\bibitem{wang2020kernelized}
H.~Wang, Y.~Wang, Z.~Zhang, X.~Fu, L.~Zhuo, M.~Xu, and M.~Wang, ``Kernelized
  multiview subspace analysis by self-weighted learning,'' \emph{IEEE
  Transactions on Multimedia}, vol.~23, pp. 3828--3840, 2020.

\bibitem{wang2018beyond}
Y.~Wang and L.~Wu, ``Beyond low-rank representations: Orthogonal clustering
  basis reconstruction with optimized graph structure for multi-view spectral
  clustering,'' \emph{Neural Networks}, vol. 103, pp. 1--8, 2018.

\bibitem{shen2015supervised}
F.~Shen, C.~Shen, W.~Liu, and H.~Tao~Shen, ``Supervised discrete hashing,'' in
  \emph{Proceedings of the IEEE conference on computer vision and pattern
  recognition}, 2015, pp. 37--45.

\bibitem{chen2020strongly}
Y.~Chen, Z.~Tian, H.~Zhang, J.~Wang, and D.~Zhang, ``Strongly constrained
  discrete hashing,'' \emph{IEEE Transactions on Image Processing}, vol.~29,
  pp. 3596--3611, 2020.

\bibitem{liu2021fddh}
X.~Liu, X.~Wang, and Y.-m. Cheung, ``Fddh: Fast discriminative discrete hashing
  for large-scale cross-modal retrieval,'' \emph{IEEE Transactions on Neural
  Networks and Learning Systems}, 2021.

\bibitem{indyk1998approximate}
P.~Indyk and R.~Motwani, ``Approximate nearest neighbors: towards removing the
  curse of dimensionality,'' in \emph{Proceedings of the thirtieth annual ACM
  symposium on Theory of computing}, 1998, pp. 604--613.

\bibitem{weiss2008spectral}
Y.~Weiss, A.~Torralba, and R.~Fergus, ``Spectral hashing,'' \emph{Advances in
  neural information processing systems}, vol.~21, 2008.

\bibitem{liu2014discrete}
W.~Liu, C.~Mu, S.~Kumar, and S.-F. Chang, ``Discrete graph hashing,''
  \emph{Advances in neural information processing systems}, vol.~27, 2014.

\bibitem{jiang2015scalable}
Q.-Y. Jiang and W.-J. Li, ``Scalable graph hashing with feature
  transformation,'' in \emph{Twenty-Fourth International Joint Conference on
  Artificial Intelligence}, 2015.

\bibitem{kumar2011co}
A.~Kumar, P.~Rai, and H.~Daume, ``Co-regularized multi-view spectral
  clustering,'' \emph{Advances in neural information processing systems},
  vol.~24, 2011.

\bibitem{8052206}
K.~Zhan, C.~Zhang, J.~Guan, and J.~Wang, ``Graph learning for multiview
  clustering,'' \emph{IEEE Transactions on Cybernetics}, vol.~48, no.~10, pp.
  2887--2895, 2018.

\bibitem{wang2016iterative}
Y.~Wang, W.~Zhang, L.~Wu, X.~Lin, M.~Fang, and S.~Pan, ``Iterative views
  agreement: an iterative low-rank based structured optimization method to
  multi-view spectral clustering,'' in \emph{Proceedings of the Twenty-Fifth
  International Joint Conference on Artificial Intelligence}, 2016, pp.
  2153--2159.

\bibitem{8662703}
H.~Wang, Y.~Yang, and B.~Liu, ``Gmc: Graph-based multi-view clustering,''
  \emph{IEEE Transactions on Knowledge and Data Engineering}, vol.~32, no.~6,
  pp. 1116--1129, 2020.

\bibitem{8778709}
H.~Xiao, Y.~Chen, and X.~Shi, ``Knowledge graph embedding based on multi-view
  clustering framework,'' \emph{IEEE Transactions on Knowledge and Data
  Engineering}, vol.~33, no.~2, pp. 585--596, 2021.

\bibitem{9492299}
S.~Shi, F.~Nie, R.~Wang, and X.~Li, ``Multi-view clustering via nonnegative and
  orthogonal graph reconstruction,'' \emph{IEEE Transactions on Neural Networks
  and Learning Systems}, pp. 1--14, 2021.

\bibitem{zhang2018binary}
Z.~Zhang, L.~Liu, F.~Shen, H.~T. Shen, and L.~Shao, ``Binary multi-view
  clustering,'' \emph{IEEE transactions on pattern analysis and machine
  intelligence}, vol.~41, no.~7, pp. 1774--1782, 2018.

\bibitem{6406693}
W.~Jin, S.~Chaki, C.~Cohen, A.~Gurfinkel, J.~Havrilla, C.~Hines, and
  P.~Narasimhan, ``Binary function clustering using semantic hashes,'' in
  \emph{2012 11th International Conference on Machine Learning and
  Applications}, vol.~1, 2012, pp. 386--391.

\bibitem{tian2020unsupervised}
Z.~Tian, H.~Zhang, Y.~Chen, and D.~Zhang, ``Unsupervised hashing based on the
  recovery of subspace structures,'' \emph{Pattern Recognition}, vol. 103, p.
  107261, 2020.

\bibitem{liu2010robust}
G.~Liu, Z.~Lin, and Y.~Yu, ``Robust subspace segmentation by low-rank
  representation,'' in \emph{Proceedings of the 27th International Conference
  on International Conference on Machine Learning}, 2010, pp. 663--670.

\bibitem{wang2017robust}
D.~Wang, Q.~Wang, and X.~Gao, ``Robust and flexible discrete hashing for
  cross-modal similarity search,'' \emph{IEEE Transactions on Circuits and
  Systems for Video Technology}, vol.~28, no.~10, pp. 2703--2715, 2017.

\bibitem{wang2021set}
W.~Wang, Y.~Shen, H.~Zhang, Y.~Yao, and L.~Liu, ``Set and rebase: determining
  the semantic graph connectivity for unsupervised cross-modal hashing,'' in
  \emph{Proceedings of the Twenty-Ninth International Conference on
  International Joint Conferences on Artificial Intelligence}, 2021, pp.
  853--859.

\bibitem{wang2021cluster}
L.~Wang, J.~Yang, M.~Zareapoor, and Z.~Zheng, ``Cluster-wise unsupervised
  hashing for cross-modal similarity search,'' \emph{Pattern Recognition}, vol.
  111, p. 107732, 2021.

\bibitem{zhang2018highly}
Z.~Zhang, L.~Liu, J.~Qin, F.~Zhu, F.~Shen, Y.~Xu, L.~Shao, and H.~T. Shen,
  ``Highly-economized multi-view binary compression for scalable image
  clustering,'' in \emph{Proceedings of the European Conference on Computer
  Vision (ECCV)}, 2018, pp. 717--732.

\bibitem{nie2016parameter}
F.~Nie, J.~Li, X.~Li \emph{et~al.}, ``Parameter-free auto-weighted multiple
  graph learning: a framework for multiview clustering and semi-supervised
  classification.'' in \emph{IJCAI}, 2016, pp. 1881--1887.

\bibitem{7876761}
C.~Hou, F.~Nie, H.~Tao, and D.~Yi, ``Multi-view unsupervised feature selection
  with adaptive similarity and view weight,'' \emph{IEEE Transactions on
  Knowledge and Data Engineering}, vol.~29, no.~9, pp. 1998--2011, 2017.

\bibitem{hartigan1979algorithm}
J.~A. Hartigan and M.~A. Wong, ``Algorithm as 136: A k-means clustering
  algorithm,'' \emph{Journal of the royal statistical society. series c
  (applied statistics)}, vol.~28, no.~1, pp. 100--108, 1979.

\bibitem{zhan2018graph}
K.~Zhan, C.~Niu, C.~Chen, F.~Nie, C.~Zhang, and Y.~Yang, ``Graph structure
  fusion for multiview clustering,'' \emph{IEEE Transactions on Knowledge and
  Data Engineering}, vol.~31, no.~10, pp. 1984--1993, 2018.

\bibitem{shi2020auto}
S.~Shi, F.~Nie, R.~Wang, and X.~Li, ``Auto-weighted multi-view clustering via
  spectral embedding,'' \emph{Neurocomputing}, vol. 399, pp. 369--379, 2020.

\bibitem{8118130}
L.~Jin, K.~Li, H.~Hu, G.-J. Qi, and J.~Tang, ``Semantic neighbor graph hashing
  for multimodal retrieval,'' \emph{IEEE Transactions on Image Processing},
  vol.~27, no.~3, pp. 1405--1417, 2018.

\bibitem{guan2019graph}
J.~Guan, Y.~Li, J.~Sun, X.~Wang, H.~Zhao, J.~Zhang, Z.~Liu, and S.~Qi,
  ``Graph-based supervised discrete image hashing,'' \emph{Journal of Visual
  Communication and Image Representation}, vol.~58, pp. 675--687, 2019.

\bibitem{xiang2019discrete}
L.~Xiang, X.~Shen, J.~Qin, and W.~Hao, ``Discrete multi-graph hashing for
  large-scale visual search,'' \emph{Neural Processing Letters}, vol.~49,
  no.~3, pp. 1055--1069, 2019.

\bibitem{fang2019unsupervised}
Y.~Fang, H.~Zhang, and Y.~Ren, ``Unsupervised cross-modal retrieval via
  multi-modal graph regularized smooth matrix factorization hashing,''
  \emph{Knowledge-Based Systems}, vol. 171, pp. 69--80, 2019.

\bibitem{li2015learning}
S.~Li and Y.~Fu, ``Learning robust and discriminative subspace with low-rank
  constraints,'' \emph{IEEE transactions on neural networks and learning
  systems}, vol.~27, no.~11, pp. 2160--2173, 2015.

\bibitem{hu2018discrete}
D.~Hu, F.~Nie, and X.~Li, ``Discrete spectral hashing for efficient similarity
  retrieval,'' \emph{IEEE Transactions on Image Processing}, vol.~28, no.~3,
  pp. 1080--1091, 2018.

\bibitem{shen2016fast}
F.~Shen, X.~Zhou, Y.~Yang, J.~Song, H.~T. Shen, and D.~Tao, ``A fast
  optimization method for general binary code learning,'' \emph{IEEE
  Transactions on Image Processing}, vol.~25, no.~12, pp. 5610--5621, 2016.

\bibitem{mallah2013plant}
C.~Mallah, J.~Cope, J.~Orwell \emph{et~al.}, ``Plant leaf classification using
  probabilistic integration of shape, texture and margin features,''
  \emph{Signal Processing, Pattern Recognition and Applications}, vol.~5,
  no.~1, pp. 45--54, 2013.

\bibitem{wang2021fast}
S.~Wang, X.~Liu, X.~Zhu, P.~Zhang, Y.~Zhang, F.~Gao, and E.~Zhu, ``Fast
  parameter-free multi-view subspace clustering with consensus anchor
  guidance,'' \emph{IEEE Transactions on Image Processing}, vol.~31, pp.
  556--568, 2021.

\bibitem{jin2013density}
Z.~Jin, C.~Li, Y.~Lin, and D.~Cai, ``Density sensitive hashing,'' \emph{IEEE
  transactions on cybernetics}, vol.~44, no.~8, pp. 1362--1371, 2013.

\bibitem{xia2015sparse}
Y.~Xia, K.~He, P.~Kohli, and J.~Sun, ``Sparse projections for high-dimensional
  binary codes,'' in \emph{Proceedings of the IEEE conference on computer
  vision and pattern recognition}, 2015, pp. 3332--3339.

\bibitem{gong2012iterative}
Y.~Gong, S.~Lazebnik, A.~Gordo, and F.~Perronnin, ``Iterative quantization: A
  procrustean approach to learning binary codes for large-scale image
  retrieval,'' \emph{IEEE transactions on pattern analysis and machine
  intelligence}, vol.~35, no.~12, pp. 2916--2929, 2012.

\bibitem{ng2001spectral}
A.~Ng, M.~Jordan, and Y.~Weiss, ``On spectral clustering: Analysis and an
  algorithm,'' \emph{Advances in neural information processing systems},
  vol.~14, 2001.

\bibitem{liu2013multi}
J.~Liu, C.~Wang, J.~Gao, and J.~Han, ``Multi-view clustering via joint
  nonnegative matrix factorization,'' in \emph{Proceedings of the 2013 SIAM
  international conference on data mining}.\hskip 1em plus 0.5em minus
  0.4em\relax SIAM, 2013, pp. 252--260.

\bibitem{nie2017multi}
F.~Nie, G.~Cai, and X.~Li, ``Multi-view clustering and semi-supervised
  classification with adaptive neighbours,'' in \emph{Thirty-first AAAI
  conference on artificial intelligence}, 2017.

\bibitem{kang2019multiple}
Z.~Kang, Z.~Guo, S.~Huang, S.~Wang, W.~Chen, Y.~Su, and Z.~Xu, ``Multiple
  partitions aligned clustering,'' \emph{arXiv preprint arXiv:1909.06008},
  2019.

\bibitem{shi2021multi}
S.~Shi, F.~Nie, R.~Wang, and X.~Li, ``Multi-view clustering via nonnegative and
  orthogonal graph reconstruction,'' \emph{IEEE Transactions on Neural Networks
  and Learning Systems}, 2021.

\end{thebibliography}

\end{document}